\documentclass[journal]{IEEEtran}
\ifCLASSINFOpdf
   \usepackage[pdftex]{graphicx}
\else
\fi

\usepackage[dvipsnames]{xcolor}
\usepackage{url}
\usepackage{bm}
\usepackage{multirow}
\usepackage{tabularx}
\usepackage{color}
\usepackage{hyperref}
\usepackage{caption}
\usepackage{soul}

\usepackage{booktabs}
\usepackage{threeparttable}
\usepackage{times}
\usepackage{epsfig}
\usepackage{graphicx}
\usepackage{amsmath}
\usepackage{amssymb}
\usepackage{comment}
\usepackage{authblk}
\usepackage{multirow}
\usepackage{tabularx}
\usepackage{balance}
\usepackage{subfigure}
\usepackage{balance}
\usepackage{algorithm}
\usepackage{xcolor}
\usepackage{color, colortbl}
\usepackage[noend]{algpseudocode}
\usepackage{cite}
\usepackage{enumitem}
\setlist{nosep}
\usepackage{diagbox}

\usepackage{pifont}

\definecolor{lightgray}{gray}{0.9}

\definecolor{mygray}{gray}{.94}

\definecolor{ggray}{RGB}{127,127,127}
\definecolor{reda}{RGB}{202,0,0}
\definecolor{redb}{RGB}{217,148,143}
\definecolor{myyellow}{RGB}{190,144,0}
\definecolor{mygreen}{RGB}{0,136,51}
\definecolor{myblue}{RGB}{0,102,204}
\begin{document}
%
\title{Joint Learning of Salient Object Detection, \\Depth Estimation and Contour Extraction }
%
%
%

\author{Xiaoqi Zhao, Youwei Pang, Lihe Zhang and Huchuan Lu
	\thanks{This work was supported by the National Key R\&D Program of China \#2018AAA0102000, the National Natural Science Foundation of China  \#62276046 and \#61876202, and the Liaoning Natural Science Foundation  \#2021-KF-12-10.\\
		X. Zhao, Y. Pang, L. Zhang and H. Lu are with the School of Information and Communication Engineering, Dalian University of Technology, Dalian, China (e-mail: zxq@mail.dlut.edu.cn; lartpang@mail.dlut.edu.cn; zhanglihe@dlut.edu.cn; lhchuan@dlut.edu.cn).}
}

\markboth{Journal of \LaTeX\ Class Files,~Vol.~14, No.~8, August~2015}%
{Shell \MakeLowercase{\textit{et al.}}: Bare Demo of IEEEtran.cls for IEEE Journals}
%



\setcounter{figure}{-2} 
\makeatletter
\g@addto@macro\@maketitle{
  \begin{figure}[H]
  \setlength{\linewidth}{\textwidth}
  \setlength{\hsize}{\textwidth}
  \centering
    \includegraphics[width=1\textwidth]{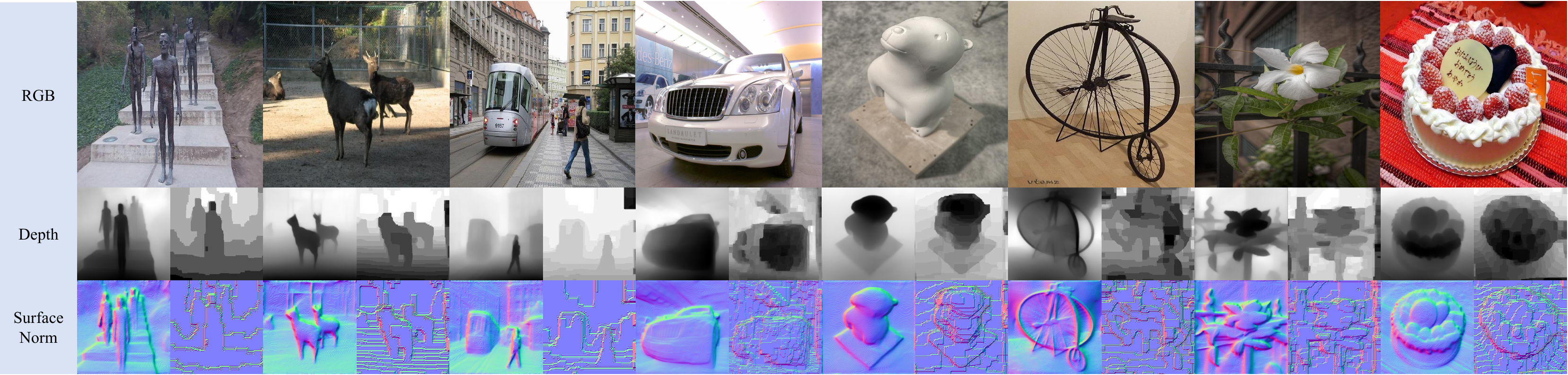}
    \caption{Comparison of the predicted depth maps (Left)  and the original ones (Right) in the STERE~\cite{STERE} dataset.}
    \label{fig:STERE_depth_surfacenorm}
  \end{figure}
  \vspace{-10mm}
}
\makeatother

\maketitle

\begin{abstract}
Benefiting from color independence, illumination invariance and location discrimination attributed by the depth map, it can provide important supplemental information for extracting salient objects in complex environments.
However, high-quality depth sensors are expensive and can not be widely applied. While general depth sensors produce the noisy and sparse depth information, which brings the depth-based networks with irreversible interference.
In this paper, we propose a novel multi-task and multi-modal filtered transformer (MMFT) network for RGB-D salient object detection (SOD). 
Specifically, we unify three complementary tasks: depth estimation, salient object detection and contour estimation.  
The multi-task mechanism promotes the model to learn the task-aware features from the auxiliary tasks. In this way, the depth information can be completed and purified. 
Moreover, we introduce a multi-modal filtered transformer (MFT) module, which equips with three modality-specific filters to generate the transformer-enhanced feature for each modality.
The proposed model works in a depth-free style during the testing phase.
Experiments show that it not only significantly surpasses the depth-based RGB-D SOD methods on multiple datasets, but also precisely predicts a high-quality depth map and salient contour at the same time. 
And, the resulted depth map can help existing RGB-D SOD methods obtain significant performance gain. The source code will be publicly available at \href{https://github.com/Xiaoqi-Zhao-DLUT/MMFT}{https://github.com/Xiaoqi-Zhao-DLUT/MMFT}.
 \end{abstract}
 

%
\IEEEpeerreviewmaketitle

\section{Introduction} \label{section:introduction}

Salient object detection (SOD) imitates the human visual attention mechanism to search attentive areas or objects from a scene and precisely segment them. As a basic computer vision task, SOD has been applied in many other tasks such as object tracking~\cite{tracking}, image caption~\cite{Imagecaption}, video object segmentation~\cite{MSAPS} and human matting~\cite{SIM}. 

In recent years, with the development of depth sensors, the RGB-D SOD task has become a very striking branch in SOD. Depth map has illumination invariance and color independence, which provides RGB image with important geometrical and discriminative information.
Most existing RGB-D SOD methods are depth-based, that is, take an RGB image and a depth map as inputs. They usually adopt a two-stream structure~\cite{BBSNet,HDFNet,UCNet,CMWNet,RD3D,SPNet} and mainly focus on how to effectively conduct the cross-modal fusion. Different from them, the DANet~\cite{DANet} designs a single stream network for efficiency. Either the two-stream methods or the single-stream one all use the depth map as input and deeply involve it into the feature learning process. As we all know, depth maps acquired by active sensors often carry different types of spatial artifacts such as Gaussian noise and hole pixels and show a checkerboard pattern due to the low distance resolution, as shown in Fig.~\ref{fig:STERE_depth_surfacenorm}. Thus, the depth-based SOD methods all inevitably face this serious interference problem. Even if a high-quality depth sensor may reduce the noise, as a single-modality device, it still needs to cooperate with other devices to handle complex scenes. 

\begin{figure*}
	\includegraphics[width=\textwidth]{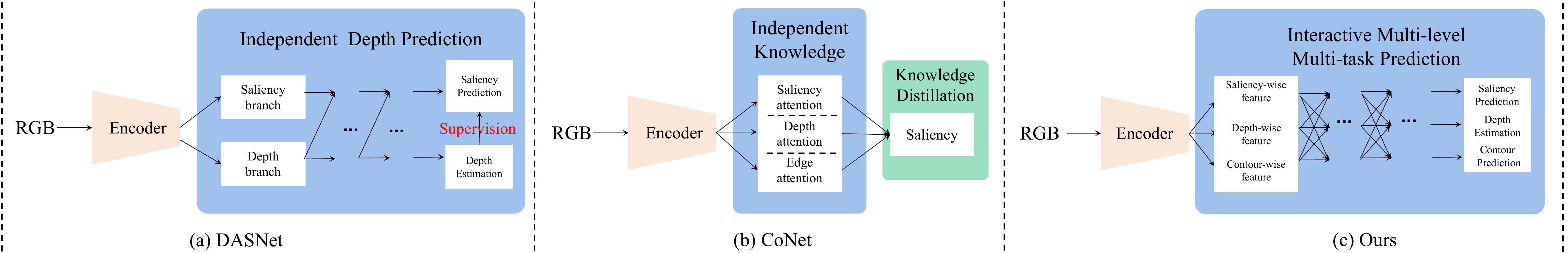}\\ 
	\centering
	\setlength{\abovecaptionskip}{-8pt}
	\caption{Comparison of three types of depth-free networks.} 
	\label{fig:three_types_depth_free}
\end{figure*} 

To this end, we rethink the RGB-D SOD problem. 
We utilize the multi-task learning (MTL) framework to implement a depth-free model, which can refine the original and coarse depth information to assist salient object detection. In particular, three complementary tasks of depth estimation, saliency and contour prediction are designed and their mutual learning achieves attention focusing and eavesdropping. 
The MTL network contains three types of modality features (depth-wise, saliency-wise, and contour-wise). How to effectively integrate them will directly affect the final predictions of these three tasks.
Motivated by the transformer~\cite{transformer}, which can capture global contextual information and establish the long-range dependence, 
we put forward a multi-modal filtered transformer (MFT) module to drive each modality stream to draw complementary information from other modalities. 
Specifically, we equip with a multi-modal transformer unit at the highest-level layer. 
Then, the proposed modality-specific filters (MSFs) are embedded in the transformer unit, which can filter out the task-aware features from the aggregated multi-modal transformer feature.
The MTL network equipped with the MFT can predict a high-quality depth map with fine and smooth distribution, as shown in Fig.~\ref{fig:STERE_depth_surfacenorm}, which is of greater benefit to saliency detection than the original depth map.

Our main contributions can be summarized as follows: 
\begin{itemize}[noitemsep,nolistsep]
	\item We develop a depth-free network based on multi-task learning framework, which can predict the high-quality depth map and saliency map even with low-quality depth supervision. Through full mutual learning among tasks, their predictions all are significantly improved. 
	\item We design a multi-modal filtered transformer module, which embeds the modality-specific filters to achieve the multi-modal feature fusion from the perspective of attention-in-attention. This design can help the task-specific stream to fully acquire useful information from auxiliary task streams.
	\item Extensive experiments on six benchmark datasets demonstrate that the proposed model performs favorably against the state-of-the-art depth-based and depth-free RGB-D SOD methods under five metrics. 
	\item The predicted depth maps have higher quality than the original ones, which can be applied to previous top-performing depth-based RGB-D SOD methods to improve their performance. 
\end{itemize}

\section{Related Work} \label{section:related_work}

\subsection{RGB-D Salient Object Detection}
According to whether the depth map is required during the testing phase, we classify existing RGB-D SOD methods into two types: depth-based and depth-free one.
\\
\textbf{Depth-based Methods.} In terms of the number of encoding streams, depth-based methods can be further divided into two-stream~\cite{PCA,CPFP,S2MA,CMWNet,PGAR,BBSNet,UCNet} and single-stream~\cite{DANet} one.  PCANet~\cite{PCA} is the first to  concentrate on the cross-modal fusion between two streams. Some later works~\cite{CPFP,ICNet,DMRA,CMWNet} leverage the depth cues to generate spatial attention or channel attention to enhance the RGB features. Sun \textit{et al.}~\cite{DSA2F} design a new NAS-based model for the heterogeneous feature fusion in RGB-D SOD and Zhou \textit{et al.}~\cite{SPNet} propose  a  novel specificity-preserving  network  to  explore  the  shared  information  between  multi-modal data. 
These two-stream designs achieve very good performance.
Recently, Zhao \textit{et al.}~\cite{DANet} combine depth map and RGB image from starting to build a single-stream network.
\\
\textbf{Depth-free Methods.} To the best of my knowledge, there are three depth-free models. Inspired by knowledge distillation, A2DELE~\cite{A2DELE} first transfers the depth knowledge from the depth stream to the RGB stream in the training phase and then only employs the RGB stream for saliency prediction when testing. DASNet~\cite{DASNet} utilizes depth information as an error-weighted map to correct the saliency prediction. As shown in Fig.~\ref{fig:three_types_depth_free} (a), it focuses on the single one task (SOD), the depth branch is just to serve the SOD branch. The most related work is CoNet~\cite{CoNet}. As shown in Fig.~\ref{fig:three_types_depth_free} (b), it first builds three independent branches to generate saliency, depth and edge features under the corresponding supervisions, and then distills all knowledge into the saliency head through multi-modal feature fusion. Different from them, as shown in Fig.~\ref{fig:three_types_depth_free} (c), we apply multi-task learning and deeply achieve cross-modal interaction to jointly train three task branches until the output. In this way, the depth branch can continuously capture useful multi-scale features from the other two branches to refine the depth information. While CoNet~\cite{CoNet} adopts the prediction-to-interaction strategy, which does not implement cross-modal interaction before predicting depth map and salient contour. Thus, its saliency 
result is very easily corrupted with the noise of the depth modality. 
\begin{figure*}
	\includegraphics[width=\textwidth]{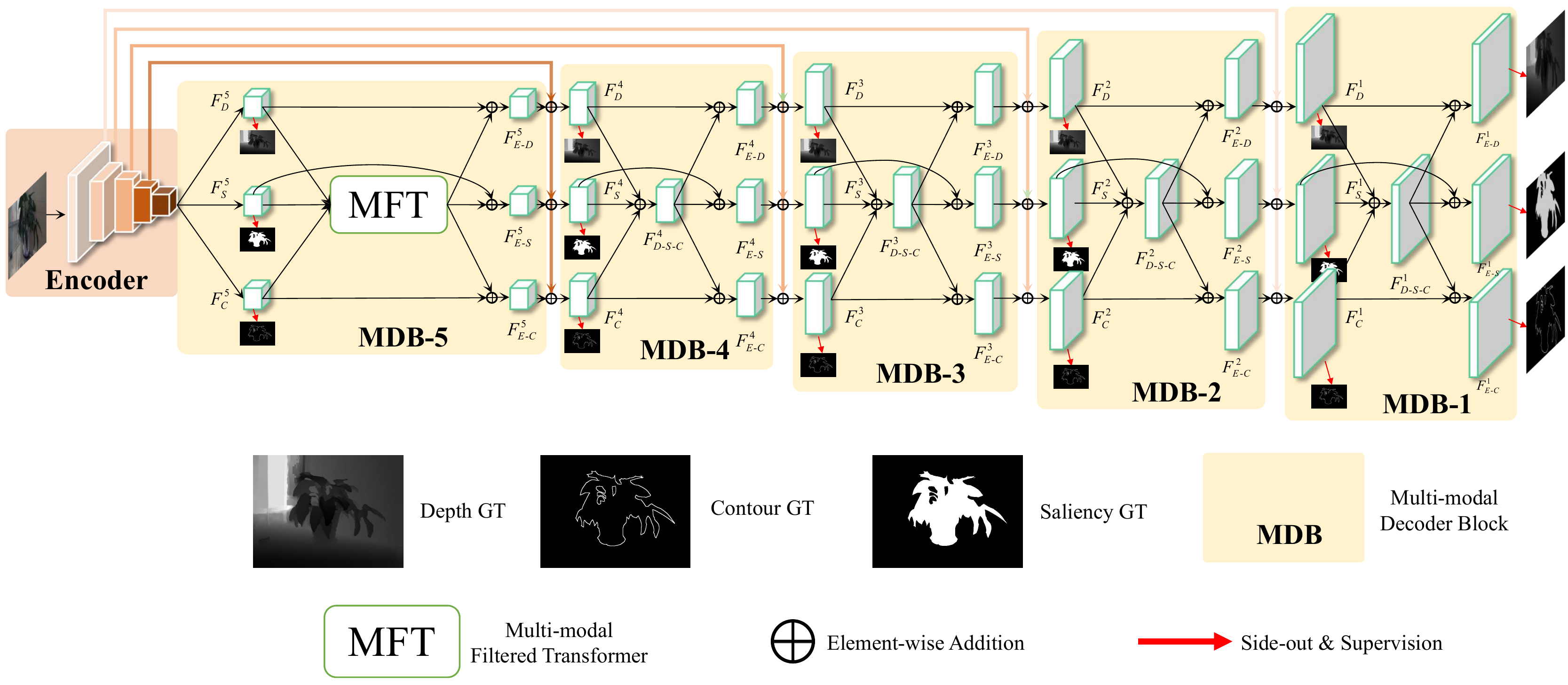}\\ 
	\centering
	\setlength{\abovecaptionskip}{-8pt}
	\caption{Network pipeline.  It consists of a encoder, five multi-modal decoder  blocks (MDBs). The multi-modal filtered transformer module is only embedded in the MDB-5. We supervise the side-out of each task branch on multiple levels.} 
	\label{fig:pipeline}
\end{figure*} 

\subsection{Multi-task Learning}
Multi-task learning (MTL) aims to solve multiple tasks by a shared model, which can improve the performance of each individual task via four implicit advantages: the implicit data augmentation, attention focusing, eavesdropping and regularization. 
In the context of deep Learning, MTL is typically done with two strategies: hard~\cite{Hard_parameter_sharing} and soft~\cite{soft_parameter_sharing} parameter sharing of hidden layers. Hard parameter sharing is the most commonly used approach in neural networks~\cite{Hard_parameter_sharing_1,Hard_parameter_sharing_2,Hard_parameter_sharing_3}, which usually shares the hidden layers among all tasks, while keeping several task-specific output layers. 
In this paper, we also adopt this strategy.
The three tasks share the multi-scale feature extractor, and the task-specific hidden layers in the decoder mutually interweave to achieve attention focusing and regularization under the supervision of ground truth, which guarantees the complementary cues to be fully mined and utilized, thereby making the three tasks complement and reinforce each other. 
\subsection{Transformer in Computer Vision}
Transformer was first proposed in ~\cite{transformer}. Due to its outstanding performance, it has revolutionized machine translation and natural language processing.  Recently, DETR~\cite{DETR} applies a transformer network for object detection, which greatly simplifies the traditional detection pipeline and shows excellent performance. And then, more and more transformer models are applied to image classification~\cite{transformer_classification}, object detection~\cite{transformer_detection}, semantic segmentation~\cite{transformer_ss}, object tracking~\cite{transformer_tracking}, video instance segmentation~\cite{transformer_vis}, etc. ViT~\cite{ViT} introduces the transformer for image recognition and models an image as a sequence of patches, which attains excellent results compared to the state-of-the-art CNN. The above works show the effectiveness of the transformer in computer vision.  However, previous works mainly focus on a single modality. How to utilize the transformer technique to achieve multi-modal fusion is what we are interested in. 

\section{The Proposed Method}
\subsection{Overall Architecture}
As shown in Fig.~\ref{fig:pipeline}, the network  consists of five encoder blocks, five multi-modal decoder blocks (MDB) and a multi-modal filtered transformer (MFT) module. The encoder-decoder architecture is based on the FPN~\cite{FPN}. The encoder is built on top of a common backbone network, e.g.,  VGG~\cite{VGG}, ResNet~\cite{Resnet} or Res2Net~\cite{Res2Net}, to perform feature extraction of  RGB image. Each MDB is used for feature communication among multiple tasks at the corresponding level. There are two reasons to equip with the MFT only on the highest-level features: (1) These features have the lowest spatial resolution, which is helpful to reduce the amount of calculations. (2) As suggested in~\cite{high_level_attention}, stand-alone attention layers can better integrate global information that usually exists in the high-level layer.

\subsection{Multi-modal Decoder Block}\label{sec:MDB}
To effectively accomplish multi-task learning, it is crucial to cooperatively utilize multi-modal information to enhance each modality branch.  As shown in Fig.~\ref{fig:pipeline}, the basic multi-modal decoder block adopts an information aggregation and distribution structure, dubbed as  \textit{Squeeze-and-Expand}. For each decoder level, we first complete the modality transfer by using the  ground truth of depth, saliency and contour to separately supervise their own feature maps. Next, we conduct  the multi-modal fusion through the \textit{Squeeze} operation. In this paper, except for the MDB-5, the other four decoder blocks directly adopt an element-wise addition layer and a convolution layer to achieve  feature fusion, as follows:
\begin{equation}\label{equ:1}
\begin{split}
F_{DSC}^{i} = Conv(F_{D}^{i} \oplus F_{S}^{i} \oplus  F_{C}^{i}), \quad i \in \left \{1, 2, 3, 4 \right \},
\end{split}
\end{equation}
where $\oplus$ is the element-wise addition and $Conv(\cdot)$ denotes the convolution layer. $F_{DSC}^{i}$ contains the depth-wise, saliency-wise and contour-wise information.  And then, we separately combine $F_{DSC}^{i}$ with each modality feature through different convolution layer, coined as the \textit{Expand} operation. 
In fact, the \textit{Squeeze-and-Expand} can be regarded as a multi-modal residual structure. 
We apply the proposed  \textit{Squeeze-and-Expand}  in each MDB to achieve the interaction between different tasks, which can effectively integrate multi-task information and then enhance the characteristics of each task. For this reason, we adopt the form of residual fusion, that is, multi-task aggregation features are used as the residual branch of each task feature to provide complementary information. Because the supervision is enforced in each MDB for each task, the supplementary features of the \textit{Squeeze-and-Expand} structure are adaptive for different tasks.  Finally, the enhanced features ($F_{enD}^{i}$, $F_{enS}^{i}$, $F_{enC}^{i}$)  combine the RGB encoder features to participate in the following multi-modal decoder block. 
\begin{figure}[t]
	\includegraphics[width=\linewidth]{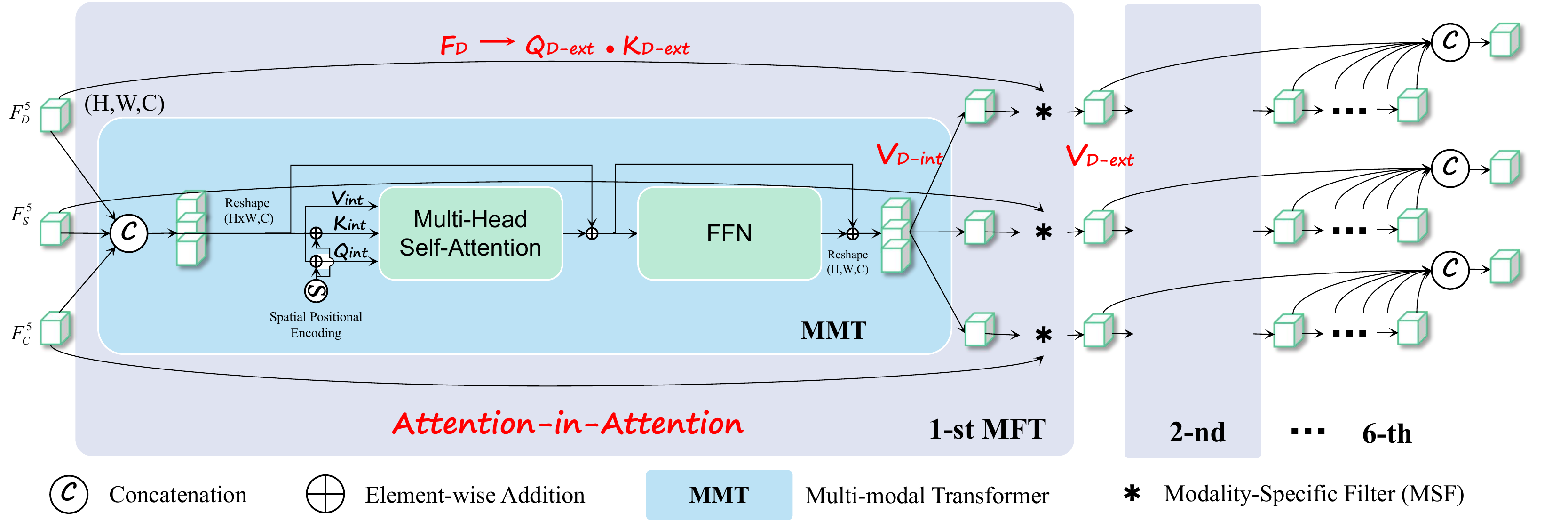}\\ 
	\centering
	\setlength{\abovecaptionskip}{-8pt}
	\caption{Illustration of the multi-modal filtered transformer module.} 
	\label{fig:mmft}
\end{figure} 
\subsection{Multi-modal Filtered Transformer}
In the MDB-5, we design a multi-modal filtered transformer (MFT) module to better perceive the global information. As shown in Fig.~\ref{fig:mmft}, it includes the multi-head self-attention, feedforward network and modality-specific filter.
\\
\textbf{Multi-modal Transformer.} 
Self-Attention (SA) is a fundamental component in standard transformer.  Given queries $q$, keys $k$ and values $v$, the attention function is written as:
\begin{equation}\label{equ:2}
\begin{split}
SA(q,k,v) = \mathcal{S}((\frac{{q}{k}^\top}{\sqrt{d_k}}){v}),
\end{split}
\end{equation}
where $d_{k}$ is the dimension of $k$ and $\mathcal{S}$ is softmax function.
The multi-head self-attention (MHSA) is formulated as:
\begin{equation}
\label{equ:3}
{MHSA}(q,k,v) = \mathcal{C}(T_{1},...,T_{nh}){\mathbf{W}^O}, 
\end{equation}
\begin{equation}
\label{equ:4}
T_{i} = SA({q}\mathbf{W}_i^q,{k}\mathbf{W}_i^k,{v}\mathbf{W}_i^v),
\end{equation}
where $\mathcal{C}$ is the concatenation operation along channel axis, $\mathbf{W}_i^Q \in \mathbb{R}^{d_m \times d_k}$, $\mathbf{W}_i^K \in \mathbb{R}^{d_m \times d_k}$, $\mathbf{W}_i^V \in \mathbb{R}^{d_m \times d_v}$, and $\mathbf{W}^O \in \mathbb{R}^{n_h\times d_v \times d_m}$ are parameter matrices. 
Please refer to the literature~\cite{transformer} for more details. 
In this work, we set $nh$ = 8, $d_{m}$ = 384 and
$d_{k}$ = $d_{v}$ = $d_{m}$/$nh$ = 12. 

Another important unit in transformer is the feedforward network (FFN) module, which is used to enhance the fitting ability of the model. It consists of two linear transformation with a ReLU in between. In the multi-modal transformer, we first aggregate and reshape $F_{D}^{5}$, $F_{S}^{5}$, $F_{C}^{5}$ $ \in \mathbb{R}^{C \times H \times W}$ as the transformer input:
\begin{equation}
\label{equ:5}
T_{in} = \mathcal{R}_{1}(\mathcal{C}(F_{D}^{5},F_{S}^{5},F_{C}^{5})), 
\end{equation}
where $\mathcal{R}_{1}$ reshapes the input feature map to $\mathbb{R}^{C \times N}$. $N = {H \times W}$ is the number of features.  Next,  $T_{in}$ is fed into MHSA and FFN in order, and then the transformer-based multi-modal fusion feature (TMFF) is generated as follows:
\begin{equation}
\label{equ:6}
MHSA_{out} = T_{in} + MHSA(T_{in} +P_{x},T_{in} +P_{x},T_{in}), 
\end{equation}
\begin{equation}
\label{equ:7}
TMFF = \mathcal{R}_{2}(MHSA_{out} + FFN(MHSA_{out})), 
\end{equation}
where $\mathcal{R}_{2}$ reshapes the input feature maps to $\mathbb{R}^{C \times H \times W}$ and $P_{x}$ is the spatial positional encoding generated in the same way as DETR~\cite{DETR} to distinguish the position information of the input feature sequence. Since the transformer has many encoder layers, it will dilute the modality-specific information to directly convey TMFF to subsequent layers. To solve this problem, we embed a modality-specific filter to select the useful features.
\begin{figure}[t]
	\includegraphics[width=\linewidth]{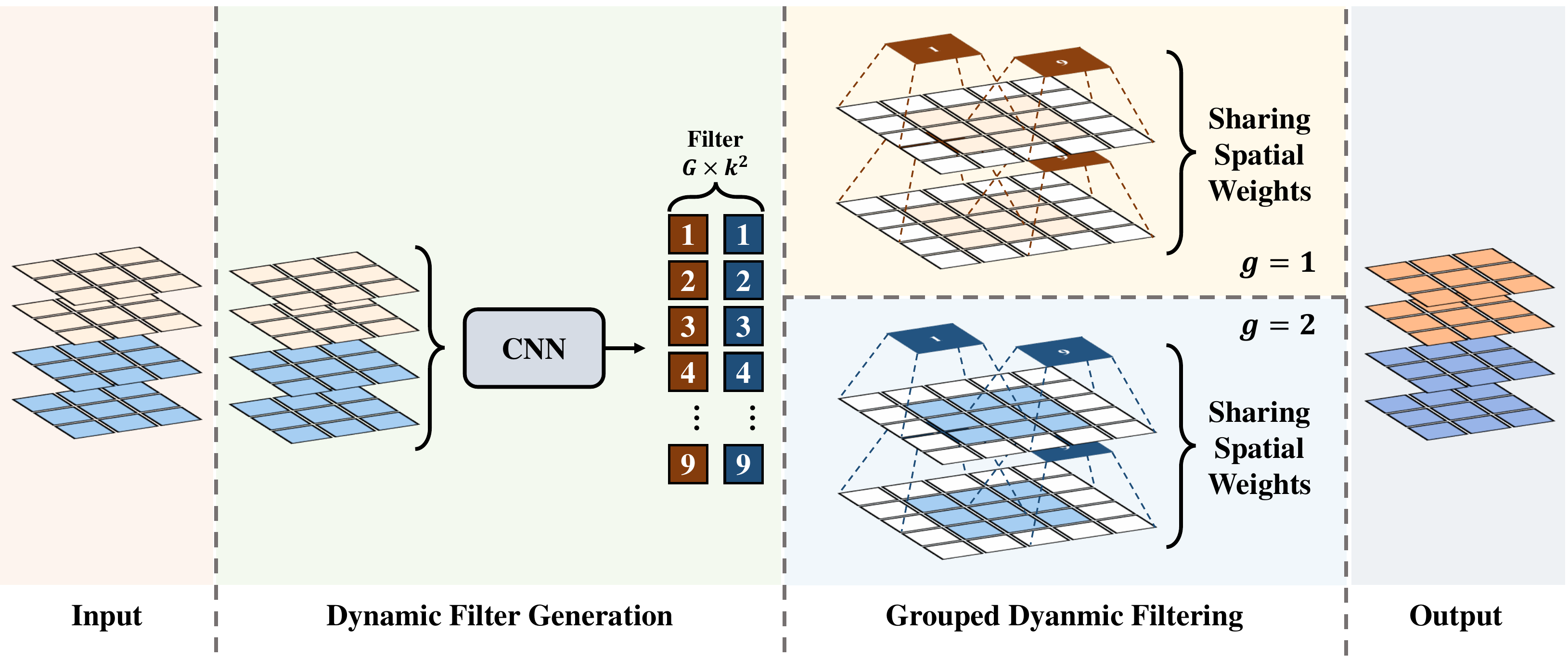}\\ 
	\centering
	\setlength{\abovecaptionskip}{-8pt}
	\caption{Detailed illustration of the modality-specific filter.} 
	\label{fig:msf}
\end{figure} 
\\
\textbf{Modality-Specific Filter.}Fig.~\ref{fig:msf} shows the dynamic filter generation and grouped dynamic filtering operation. We follow the spatially-unshared and intragroup-shared principles to design the filter.
On the one hand, the differences among the predictions of three tasks are mainly reflected in the location properties of the pixel level. 
Therefore, we adopt the dynamic kernel to capture the position-specific information, that is, the filter is spatially unshared.
On the other hand, the high-level features have rich semantic information in different channels. In order to maintain semantic diversity, we adopt the channel-unshared strategy. 
However, to save the number of parameters, we refer to the idea of group convolution~\cite{ResNext} and adopt a grouping mechanism to achieve the intragroup-shared but intergroup-unshared filtering. 
The filtering process can be formulated as:
\begin{equation}
\label{equ:8}
Y_{c_{i},h,w} = \sum_{u,v}F_{\lceil \frac{c_{i}}{C_{g}} \rceil,h,w,u+\lfloor \frac{K_{h}}{2} \rfloor,v+\lfloor \frac{K_{w}}{2} \rfloor}X_{c_{i},h+u,w+v},
\end{equation}
where $X\in \mathbb{R}^{C_{i} \times H \times W}$ and $Y\in \mathbb{R}^{C_{o} \times H \times W}$ are the input and output features, respectively.  $F\in \mathbb{R}^{G \times H \times W \times K_{h} \times K_{w}}$ denotes the dynamic filters, which are yielded by feeding $X$ to two $1$ $\times$ $1$ vanilla convolutional layers. $K_{h} \times K_{w}$ represents the size of the convolution kernel. In this work, we use $3 \times 3$ kernel, and set $G=8$ and $C_{g}=C_{i}//G$. 
For convenience, the offset values relative to the center of the convolution kernel are indexed as $(u,v)\in\Delta_{K}=[-\lfloor \frac{K_{h}}{2}\rfloor,...,\lfloor \frac{K_{h}}{2}\rfloor]\times[-\lfloor \frac{K_{w}}{2}\rfloor,...,\lfloor \frac{K_{w}}{2}\rfloor]\in\mathbb{Z}^{2}$, where $\times$ is the Cartesian product and  $\Delta_{K}$ forms a set of two-dimensional coordinate offset. 
\\
\textbf{Attention-in-Attention.} 
In fact, an \textit{Attention-in-Attention} mechanism is implemented by combining the multi-modal transformer and the modality-specific filter. 
The self-attention establishes the intrinsic correlation of queries $Q$, keys $K$ and values $V$, which provides a tight information exchange site for an isolated source of the input feature $F$. The frequently cascaded self-attention may gradually weaken the representation capability of features due to the possible accumulation of modality mismatch.
For this reason, we build an additional attention barrier outside the internal attention, which takes the output $V_{m-int},m\in\{D, S, C\}$ of the internal attention as the input of the external attention and then utilizes  $F_m$ to generate attention weight $Q_{m-ext} \cdot K_{m-ext}$. The external attention establishes the secondary relationship among $Q$, $K$ and  $V$ through the filtering process. Thus, the internal $V_{m-int}$ is filtered and its enhanced part $V_{m-ext}$ is obtained under the bridging guidance of the external attention. The external-internal attention is actually an attention-in-attention structure. The correspondence between the multi-modal filtered transformer and the \textit{Attention-in-Attention} is shown in Fig.~\ref{fig:mmft}.

\subsection{Supervision}
The total training loss can be written as:
\begin{equation}\label{equ:9}
\begin{split}
L= L_{d}+L_{s}+L_{c}.
\end{split}
\end{equation}
For the depth estimation, we follow the related works~\cite{depth3,depth4,depth5} to adopt the combination loss of the L1 and SSIM~\cite{SSIM}.
For the saliency prediction, we use the weighted IoU loss and binary cross entropy (BCE) loss which have been widely adopted in segmentation tasks. We use the same definitions as in~\cite{SPNet,F3Net,PraNet,MSNet}.
For the contour prediction, we only adopt the BCE loss.
\begin{align}
L_{d}= & \sum_{n=1}^{N}(l_{ssim}^{(n)}+l_{1}^{(n)})\\
L_{s}= &\sum_{n=1}^{N}(l_{bce}^{w(n)} + l_{iou}^{w(n)})
\\
L_{c}= &\sum_{n=1}^{N} l_{bce}^{(n)}
\end{align}

As described in Sec.~\ref{sec:MDB}, our model is deeply supervised on five groups of outputs, i.e. $N$ = $5$. 
Besides, for the salient contour estimation, its ground truth ($G_{c}$) is calculated based on the ground-truth saliency map ($G_{s}$). Specifically, we employ the morphological dilation and erosion as follows:
\begin{equation}\label{equ:5}
\begin{split}
G_{c}= D_{m}(G_{s})-E_{m}(G_{s}),
\end{split}
\end{equation}
where $D(\cdot)$ and $E(\cdot)$ are the dilation and erosion operations, respectively. $m$ denotes a full one filter of size $m\times m$ for erosion and expansion, which is set to $3$ in this paper. 

\section{Experiments}
\subsection{Datasets}\label{sec:datasets_sec}
We evaluate the proposed model on six public RGB-D SOD datasets, which are \emph{DUTLF-D}~\cite{DMRA}, \emph{STERE}~\cite{STERE}, \emph{NJUD}~\cite{NJU2000}, \emph{NLPR}~\cite{early_fusion_1}, \emph{SIP}~\cite{SIP} and \emph{LFSD}~\cite{LFSD}. For comprehensive and fair evaluation, we follow the setting as most previous methods~\cite{SPNet,HDFNet,BBSNet,CMWNet,JLDCF,CoNet,A2DELE}. On the DUTLF-D, we  use  $800$  samples  for  training  and  $400$ samples for  testing. For  the  other  five  datasets,  we use  $1,485$ samples from the NJUD and $700$ samples from the NLPR as the training set and the remaining samples in these datasets are used for testing. Besides, we quantitatively compare the predicted depth map of our method and  CoNet~\cite{CoNet} and other two state-of-the-art depth estimation methods~\cite{TL-Depth,LAP-Depth} on the SIP. This dataset consists of high-resolution RGB color images and depth maps which are designed to contain multiple salient persons per image. The  high-quality camera makes sure that the depth map from the SIP has less noise. 

\subsection{Evaluation Metrics}
For the salient object detection task, we adopt several widely used metrics for quantitative evaluation: F-measure ($F_{\beta}^{max}$)~\cite{colorcontrast_Fm}, the weighted F-measure ($F_{\beta}^{w}$)~\cite{Fwb}, mean absolute error ($\mathcal{M}$), the recently released S-measure ($S_{m}$)~\cite{S-m} and E-measure ($E_{m}$)~\cite{Em}. Besides, we use the Area Under Curve (AUC) as the complementary binary classification evaluation metric. The lower value is better for the $\mathcal{M}$ and the higher is better for others. For the depth estimation task, we use following metrics used  by previous works: Root Mean Square Error (RMSE, RMSE (log)), absolute relative error (AbsRel), squared relative error (SqRel) and depth accuracy at various thresholds $1.25$, $1.25^{2}$ and $1.25^{3}$.
\subsection{Implementation Details}\label{sec:implementationdetails}
Our model is implemented based on the Pytorch and trained on one RTX 3090 GPU for $50$ epochs with mini-batch size $12$.  The input resolutions of RGB images are resized to $352\times352$. We adopt some image augmentation techniques to avoid overfitting, including random flipping,  rotating, and border clipping. For the optimizer, we use the Adam~\cite{Adam}. For the learning rate, initial learning rate  is  set  to $0.0001$. We adopt the ``step'' learning rate decay policy, and set the decay size as $30$ and decay rate as $0.9$. Our code will be made publicly available soon.
\subsection{Comparisons with State-of-the-art}
\subsubsection{Quantitative Evaluation}
~\\
\noindent{\textbf{\textit{Saliency Evaluation.}}}
Since previous RGB-D SOD methods adopt VGG-16, ResNet-50, Res2Net-50 or ResNet-101 as the backbone. For a fair comparison, we separately evaluate the MMFT with different backbones.
Tab.~\ref{tab:RGBDSOD_performance} shows performance comparisons in terms of five metrics. We divide these models into depth-based and depth-free. Our  method  outperforms  all  of  the  competitors and obtains the best performance in terms of all metrics on the DUTLF-D, STERE and SIP datasets. It is worth noting that our model achieves significant improvement against the second best depth-free (CoNet~\cite{CoNet}) and depth-based (SPNet~\cite{SPNet}) models.
In addition, the LFSD dataset has only 100 samples, which was originally used for light field saliency research. Most of samples have small variation in depth direction and have large-size foreground, which easily results in that the effect of high-quality depth map is weakened, thereby suppressing the performance of multi-task learning. 
Tab.~\ref{tab:para_flops} lists the \textit{\textbf{parameters and FLOPs}} of different methods with superior performance in Tab.~\ref{tab:RGBDSOD_performance}. It can be seen that MMFTs still have advantages against most state-of-the-art models with different backbones under the FLOPs metric. Our method achieves a good balance between accuracy and efficiency while accomplishing multiple tasks.

Besides, we also conduct an extension for RGB SOD task. We use the MMFT, which is trained on RGB-D SOD datasets, to predict depth maps for the images in RGB SOD datasets, and then follow most RGB SOD methods~\cite{SGL-KRN,Auto-MSFNet,SAMNet,VST,F3Net,GCPA,MINet,ITSD,GateNet} to re-train the MMFT on the DUTS-TR~\cite{DUTS} dataset for RGB SOD task. That is, the predicted depth map is used to supervise the depth branch. Tab.~\ref{tab:rgbsod_performance} shows the results in terms of five metrics on five RGB SOD datasets. Our method performs favorably against these competitors. Especially, it achieves the best performance on the challenging DUT-OMORN~\cite{DUT-OMRON} dataset. We note that the proposed method performs slightly bad on the HKU-IS dataset. The reason is possibly as follows: This dataset focuses on multi-object saliency segmentation. Its scenes are usually not complicated and the contrast between the foreground and background is obvious. Thus, the auxiliary effect of depth information for saliency prediction is limited. We think that applying some multi-scale designs such as ASPP or Dense-ASPP can improve the performance on this dataset.
\\
\textbf{\textit{Saliency Contour Evaluation.}} We evaluate salient contour in terms of two main metrics: $F_{\beta}^{max}$ and $\mathcal{M}$ in Tab.~\ref{tab:Salient_contour_performance}. It can be seen that our method surpasses these competitors on all five datasets. Their contours are extracted from the corresponding saliency maps by morphological processing.
\begin{table*}
	\scriptsize
	\setlength{\abovecaptionskip}{2pt}
	\caption{Quantitative comparison of different RGB-D SOD methods. $\uparrow$ and $ \downarrow$ indicate that the larger scores and the smaller ones  are better, respectively. The best three scores are highlighted in {\color{reda} \textbf{red}}, {\color{mygreen} \textbf{green}} and {\color{myblue} \textbf{blue}}. 
		The subscript in each model name is the publication year.  The corresponding used backbone network is listed below each model name.}
	\renewcommand\tabcolsep{1pt} 
	\renewcommand\arraystretch{1.3}
	\centering
	\resizebox{\textwidth}{!}  
	{
		\begin{tabular}{cr||cccccccccccccccc||ccc>{\columncolor{mygray}}c>{\columncolor{mygray}}c>{\columncolor{mygray}}c>{\columncolor{mygray}}c}
			\toprule[2pt]
			\multicolumn{2}{l||}{} & \multicolumn{16}{c||}{\textbf{{Depth-based Models}}}  & \multicolumn{7}{c}{\textbf{{Depth-free Models}}} \\
			\hline
			\multicolumn{2}{c||}{\multirow{3}{*}{Metric}}   &{S2MA$_{20}$} & {CMWNet$_{20}$} & {BBSNet$_{20}$} & {BBSNet$_{20}$} &{JL-DCF$_{20}$}  & {UCNet$_{20}$} & {PGAR$_{20}$} &{HDFNet$_{20}$}
			&{HDFNet$_{20}$}
			&{CDNet$_{21}$} &{DFMNet$_{21}$} &{DSA2F$_{21}$} &{DCF$_{21}$}&{HAINet$_{21}$} &   {RD3D$_{21}$} &   {SPNet$_{21}$} & {A2DELE$_{20}$}&{DASNet$_{20}$}&{CoNet$_{20}$} & {MMFT} &{MMFT} & {MMFT} & {MMFT} 
			\\
			\multicolumn{2}{l||}{}   & ~\cite{S2MA}&~\cite{CMWNet}&~\cite{BBSNet}&~\cite{BBSNet}&~\cite{JLDCF}&~\cite{UCNet}&~\cite{PGAR}&~\cite{HDFNet}&~\cite{HDFNet}&~\cite{CDNet}&~\cite{DFMNet}&~\cite{DSA2F}&~\cite{DCF}&~\cite{HAINet}&~\cite{RD3D}&~\cite{SPNet}&~\cite{A2DELE}&~\cite{DASNet}&~\cite{CoNet}&\cellcolor{mygray}&\cellcolor{mygray}&\cellcolor{mygray}&\cellcolor{mygray}
			\\
			\multicolumn{2}{l||}{}   & VGG-16  & VGG-16& VGG-16& ResNet-50&ResNet-101& VGG-16& VGG-16&  VGG-16& ResNet-50& VGG-16 &ResNet-34& VGG-19& ResNet-50& VGG-16& ResNet-50& Res2Net-50&  VGG-16& ResNet-50&ResNet-101& VGG-16& ResNet-50& Res2Net-50& ResNet-101
			\\
			\hline
			\hline
			\multirow{5}{*}{\emph{\rotatebox{90}{DUTLF-D}}}      
			&$F_{\beta}^{max}\uparrow$   
			& 0.909 
			& 0.905 
			& -
			& -
			& 0.924 
			& -
			& 0.938 
			& 0.926 
			& 0.930 
			& 0.901 
			& -
			& 0.938 
			& 0.941 
			& 0.932 
			& 0.946 
			& -
			& 0.907 
			& -
			& 0.935 
			& 0.950 
			& \color{myblue} \textbf{0.955}
			& \color{reda} \textbf{0.958} 
			& \color{mygreen} \textbf{0.956}
			
			\\
			&$F_{\beta}^{w}\uparrow$    
			&  0.862 
			&0.831 
			&-
			&-
			&0.863 
			&-
			&0.889 
			&0.865 
			&0.864 
			&0.838 
			&-
			&0.908 
			&0.909 
			&0.883 
			&0.909 
			&-
			&0.864 
			& -
			&0.891 
			&0.909 
			& \color{myblue} \textbf{0.920} 
			& \color{reda} \textbf{0.926} 
			& \color{mygreen} \textbf{0.924} 
			\\
			& $S_m\uparrow$  
			& 0.903 
			&0.887 
			&-
			&-
			&0.905 
			&-
			&0.920 
			&0.905 
			&0.907 
			&0.886 
			&-
			&0.921 
			&0.924 
			&0.909 
			&0.931 
			&-
			&0.886 
			& -
			&0.919 
			&0.934 
			& \color{myblue} \textbf{0.941} 
			& \color{mygreen} \textbf{0.942} 
			& \color{reda} \textbf{0.943} 
			
			\\
			& $E_m\uparrow$    &
			0.921 
			&0.922 
			&-
			&-
			&0.938 
			&-
			&0.950 
			&0.938 
			&0.938 
			&0.917 
			&-
			&0.956 
			&0.957 
			&0.939 
			&0.957 
			&-
			&0.929 
			& -
			&0.952 
			& \color{myblue} \textbf{0.959} 
			& \color{mygreen} \textbf{0.967} 
			& \color{mygreen} \textbf{0.967} 
			& \color{reda} \textbf{0.968} 
			\\
			&$\mathcal{M}\downarrow$  &   0.044 
			& 0.056 
			& -
			& -
			& 0.043 
			& -
			& 0.035 
			& 0.040 
			& 0.041 
			& 0.051 
			& -
			& 0.031 
			& 0.030 
			& 0.038 
			& 0.031 
			& -
			& 0.043 
			& -
			& 0.033 
			& 0.029 
			& \color{myblue} \textbf{0.027} 
			& \color{mygreen} \textbf{0.025} 
			& \color{reda} \textbf{0.024} 
			\\
				&$AUC\uparrow$ 
&.9721
&.9749
&-
&-
&.9835
&-
&.9802
&\color{mygreen} \textbf{.9892}
&\color{myblue} \textbf{.9884}
&.9567
&-
&.9662
&.9767
&.9777
&.9826
&-
&.9371
&-
&.9817
&.9879
&\color{reda} \textbf{.9893}
&.9856
&.9882

			\\
			\hline
			\multirow{5}{*}{\emph{\rotatebox{90}{STERE}}}      
			&$F_{\beta}^{max}\uparrow$   &   
			0.895 
			& 0.911 
			& 0.901 
			& 0.919 
			& 0.913 
			& 0.908 
			& 0.911 
			& 0.918 
			& 0.910 
			& 0.909 
			& 0.914 
			& 0.910 
			& 0.915 
			& 0.919 
			& 0.917 
			& 0.915 
			& 0.892 
			& 0.915 
			& 0.909 
			& \color{myblue} \textbf{0.931} 
			&  \color{mygreen} \textbf{0.933 }
			&  \color{reda} \textbf{0.935} 
			&  \color{blue} \textbf{0.931} 
			
			\\
			&$F_{\beta}^{w}\uparrow$    &   
			0.825 
			&0.847 
			&0.838 
			&0.858 
			&0.857 
			&0.867 
			&0.856 
			&0.863 
			&0.853 
			&0.855 
			&0.860 
			&0.869 
			&0.873 
			&0.871 
			&0.871 
			&0.873 
			&0.846 
			& 0.872
			&0.866 
			& \color{myblue} \textbf{0.886} 
			& \color{mygreen} \textbf{0.889} 
			& \color{reda} \textbf{0.894} 
			&  \color{mygreen} \textbf{0.889} 
			\\
			& $S_m\uparrow$  &  
			0.890 
			&0.905 
			&0.896 
			&0.908 
			&0.903 
			&0.903 
			&0.907 
			&0.906 
			&0.900 
			&0.903 
			&0.908 
			&0.897 
			&0.905 
			&0.909 
			&0.911 
			&0.907 
			&0.878 
			& 0.910
			&0.905 
			& \color{myblue} \textbf{0.923} 
			& \color{reda} \textbf{0.926} 
			& \color{mygreen} \textbf{0.925} 
			&0.921

			\\
			& $E_m\uparrow$    &  
			0.926 
			& 0.930 
			& 0.928 
			& 0.941 
			& 0.937 
			& 0.942 
			& 0.937 
			& 0.937 
			& 0.931 
			& 0.938 
			& 0.939 
			& 0.942 
			& 0.943 
			& 0.938 
			& 0.944 
			& 0.942 
			& 0.928 
			& 0.941
			& 0.941 
			& \color{myblue} \textbf{0.949} 
			& 0.947 
			&  \color{reda} \textbf{0.953} 
			&  \color{mygreen} \textbf{0.952} 
			
			\\
			&$\mathcal{M}\downarrow$  &   0.051 
			&0.043 
			&0.046 
			&0.041 
			&0.040 
			&0.039 
			&0.041 
			&0.039 
			&0.041 
			&0.041 
			&0.040 
			&0.039 
			&0.037 
			&0.038 
			&0.037 
			&0.037 
			&0.045 
			& 0.037
			&0.037 
			& \color{myblue} \textbf{0.033} 
			& \color{mygreen} \textbf{0.032} 
			& \color{reda} \textbf{0.031} 
			& \color{myblue} \textbf{0.033} 
   
			\\
   				&$AUC\uparrow$ 
&.9798
&.9846
&.9785
&.9841
&.9846
&.9762
&.9804
&\color{myblue} \textbf{.9870}
&.9854
&.9742
&.9849
&.9567
&.9740
&.9849
&.9812
&.9806
&.9355
&.9832
&.9775
&\color{mygreen} \textbf{.9873}
&\color{reda} \textbf{.9884}
&.9818
&.9820
			\\
			\hline
			\multirow{5}{*}{\emph{\rotatebox{90}{NLPR}}}      
			&$F_{\beta}^{max}\uparrow$    &   0.910 
			&0.913 
			&0.921 
			&0.927 
			&0.925 
			&0.916 
			&0.925 
			&0.917 
			&0.927 
			&0.925 
			&0.919 
			&0.916 
			&0.917 
			&0.917 
			&0.927 
			&0.926 
			&0.898 
			& 0.929
			&0.898 
			& \color{myblue} \textbf{0.931} 
			&\color{mygreen} \textbf{0.935} 
			& \color{myblue} \textbf{0.931} 
			&\color{reda} \textbf{0.937} 
			\\
			&$F_{\beta}^{w}\uparrow$    &   
			0.852 
			&0.856 
			&0.871 
			&0.879 
			&0.882 
			&0.878 
			&0.881 
			&0.869 
			&0.882 
			&0.882 
			&0.877 
			&0.881 
			&0.886 
			&0.881 
			&0.889 
			&\color{mygreen} \textbf{0.896} 
			&0.857 
			& \color{myblue} \textbf{0.895}
			&0.842 
			&0.885 
			& {0.894} 
			& {0.893} 
			& \color{reda} \textbf{0.900} 
			\\
			& $S_m\uparrow  $         &  
			0.915 
			&0.917 
			&0.923 
			&0.930 
			&0.925 
			&0.920 
			&0.930 
			&0.916 
			&0.923 
			&0.930 
			&0.925 
			&0.918 
			&0.921 
			&0.921 
			&0.929 
			&0.927 
			&0.898 
			&0.929
			&0.908 
			&0.927 
			& \color{mygreen} \textbf{0.937} 
			& \color{myblue} \textbf{0.933} 
			& \color{reda} \textbf{0.938}
			\\
			& $E_m\uparrow$  &  
			0.942 
			&0.941 
			&0.948 
			&0.954 
			&0.955 
			&0.955 
			&0.955 
			&0.948 
			&0.957 
			&0.954 
			&0.954 
			&0.952 
			&0.956 
			&0.952 
			& \color{myblue} \textbf{0.959} 
			& \color{myblue} \textbf{0.959} 
			&0.945 
			&\color{mygreen} \textbf{0.961}
			&0.934 
			&0.952 
			& {0.958} 
			&{0.958} 
			& \color{reda} \textbf{0.963} 
			\\
			&$\mathcal{M}\downarrow$    & 
			0.030 
			&0.029 
			&0.026 
			&0.023 
			&0.022 
			&0.025 
			&0.024 
			&0.027 
			&0.023 
			&0.024 
			&0.024 
			&0.024 
			&0.023 
			&0.025 
			& \color{myblue} \textbf{0.022} 
			& \color{mygreen} \textbf{0.021} 
			&0.029 
			& \color{mygreen} \textbf{0.021} 
			&0.031 
			&0.024 
			& \color{mygreen} \textbf{0.021} 
			&\color{myblue} \textbf{0.022} 
			& \color{reda} \textbf{0.020} 
   			\\
   				&$AUC\uparrow$ 
&.9829
&.9866
&.9838
&.9882
&.\color{myblue} \textbf{9889}
&.9832
&.9878
&.9861
&.9888
&.9854
&\color{mygreen} \textbf{.9894}
&.9712
&.9770
&.9842
&.9846
&.9833
&.9447
&.9847
&.9826
&.9878
&\color{reda} \textbf{.9915}
&.9847
&.9867

			\\
			\hline
			\multirow{5}{*}{\emph{\rotatebox{90}{NJUD}}}      
			&$F_{\beta}^{max}\uparrow$   &   
			0.899 
			&0.913 
			&0.926 
			&0.931 
			&0.912 
			&0.908 
			&0.918 
			&0.924 
			&0.922 
			&0.919 
			&0.921 
			&0.917 
			&0.917 
			&0.920 
			&0.923 
			&0.935 
			&0.890 
			&0.911
			&0.902 
			&\color{myblue} \textbf{0.939} 
			&0.938 
			& \color{mygreen} \textbf{0.940} 
			& \color{reda} \textbf{0.945} 
			\\
			&$F_{\beta}^{w}\uparrow$     & 
			0.842 
			&0.857 
			&0.878 
			&0.884 
			&0.869 
			&0.868 
			&0.872 
			&0.881 
			&0.877 
			&0.878 
			&0.879 
			&0.883 
			&0.878 
			&0.879 
			&0.886 
			& \color{mygreen} \textbf{0.906} 
			&0.843 
			&0.872
			&0.850 
			&0.900 
			&0.900 
			& \color{myblue} \textbf{0.905} 
			& \color{reda} \textbf{0.909} 
			\\
			& $S_m\uparrow$        &   
			0.894 
			&0.903 
			&0.916 
			&0.921 
			&0.902 
			&0.897 
			&0.909 
			&0.911 
			&0.908 
			&0.913 
			&0.912 
			&0.904 
			&0.903 
			&0.909 
			&0.916 
			&0.924 
			&0.871 
			&0.902
			&0.895 
			& \color{mygreen} \textbf{0.928} 
			& \color{mygreen} \textbf{0.928} 
			& \color{myblue} \textbf{0.927} 
			& \color{reda} \textbf{0.930} 
			
			\\
			& $E_m\uparrow$     &   
			0.917 
			&0.923 
			&0.937 
			&0.942 
			&0.935 
			&0.934 
			&0.935 
			&0.934 
			&0.932 
			&0.940 
			&0.937 
			&0.937 
			&0.941 
			&0.931 
			&0.942 
			& \color{mygreen} \textbf{0.953} 
			&0.916 
			&0.936
			&0.924 
			&0.944 
			&0.944 
			& \color{myblue} \textbf{0.950} 
			& \color{reda} \textbf{0.954} 
			\\
			&$\mathcal{M}\downarrow$  &  
			0.053 
			&0.046 
			&0.039 
			&0.035 
			&0.041 
			&0.043 
			&0.042 
			&0.037 
			&0.038 
			&0.038 
			&0.039 
			&0.039 
			&0.038 
			&0.038 
			&0.037 
			& \color{reda} \textbf{0.029} 
			&0.047 
			&0.042 
			&0.046 
			&0.033 
			&0.033 
			& \color{myblue} \textbf{0.031} 
			& \color{mygreen} \textbf{0.030} 
      			\\
   				&$AUC\uparrow$ 
&.9745
&.9821
&.9828
&.\color{myblue} \textbf{9849}
&.9812
&.9700
&.9765
&.9843
&.9835
&.9766
&.9845
&.9619
&.9660
&.9810
&.9791
&.9814
&.9296
&.9739
&.9755
&\color{reda} \textbf{.9879}
&\color{mygreen} \textbf{.9872}
&.9825
&.9832

			\\
			\hline
			\multirow{5}{*}{\emph{\rotatebox{90}{SIP}}}      
			&$F_{\beta}^{max}\uparrow$   &   0.891 
			&0.890 
			&0.892 
			&0.902 
			&0.904 
			&0.896 
			&0.893 
			&0.904 
			&0.910 
			&0.888 
			&0.904 
			&0.891 
			&0.900 
			&0.916 
			&0.906 
			&0.916 
			&0.855 
			&0.900
			&0.883 
			& \color{mygreen} \textbf{0.935} 
			&0.929 
			& \color{myblue} \textbf{0.931} 
			& \color{reda} \textbf{0.937} 
			\\
			&$F_{\beta}^{w}\uparrow$   &   0.819 
			&0.811 
			&0.820 
			&0.830 
			&0.844 
			&0.836 
			&0.822 
			&0.835 
			&0.848 
			&0.812 
			&0.842 
			&0.829 
			&0.841 
			&0.854 
			&0.845 
			&0.868 
			&0.780 
			&0.836
			&0.803 
			& \color{myblue} \textbf{0.878} 
			&0.875 
			& \color{mygreen} \textbf{0.881} 
			& \color{reda} \textbf{0.893} 
			\\
			& $S_m\uparrow$        &   0.872 
			&0.867 
			&0.874 
			&0.879 
			&0.880 
			&0.875 
			&0.876 
			&0.878 
			&0.886 
			&0.862 
			&0.885 
			&0.862 
			&0.873 
			&0.886 
			&0.885 
			&0.894 
			&0.828 
			&0.877 
			&0.858 
			& \color{myblue} \textbf{0.909} 
			& \color{mygreen} \textbf{0.911} 
			& \color{myblue} \textbf{0.909} 
			& \color{reda} \textbf{0.917} 
			\\
			& $E_m\uparrow$     &  
			0.913 
			&0.909 
			&0.912 
			&0.917 
			&0.923 
			&0.918 
			&0.912 
			&0.921 
			&0.925 
			&0.905 
			&0.925 
			&0.911 
			&0.921 
			&0.925 
			&0.924 
			&0.931 
			&0.890 
			&0.923
			&0.909 
			& \color{myblue} \textbf{0.942} 
			&0.939 
			& \color{mygreen} \textbf{0.943} 
			& \color{reda} \textbf{0.950} 
			\\
			&$\mathcal{M}\downarrow$  &   0.057 
			&0.062 
			&0.056 
			&0.055 
			&0.049 
			&0.051 
			&0.055 
			&0.050 
			&0.047 
			&0.060 
			&0.049 
			&0.057 
			&0.052 
			&0.048 
			&0.048 
			&0.043 
			&0.070 
			&0.051
			&0.063 
			& \color{myblue} \textbf{0.039} 
			& \color{mygreen} \textbf{0.038} 
			& \color{mygreen} \textbf{0.038} 
			& \color{reda} \textbf{0.034} 
         			\\
   				&$AUC\uparrow$ 
&.9585
&.9524
&.9539
&.9555
&.9575
&.9528
&.9610
&.9660
&.9676
&.9361
&.9667
&.9328
&.9391
&.9607
&.9550
&.9609
&.8974
&.9625
&.9485
&\color{reda} \textbf{.9751}
&\color{myblue} \textbf{.9749}
&.9682
&\color{mygreen} \textbf{.9750}

			\\
			\hline
			\multirow{5}{*}{\emph{\rotatebox{90}{LFSD}}}      
			&$F_{\beta}^{max}\uparrow$   
			& 0.862 
			& \color{mygreen} \textbf{0.900} 
			&0.862 
			&0.879 
			&0.887 
			&0.878 
			&0.874 
			&0.860 
			&0.883 
			&0.879 
			&0.888 
			& \color{reda} \textbf{0.903} 
			&0.878 
			&0.877 
			&0.879 
			&0.881 
			&0.858
			&0.845 
			&0.877 
			&0.892 
			&0.895 
			& \color{myblue} \textbf{0.897} 
			&0.884

			\\
			&$F_{\beta}^{w}\uparrow$    
			& 0.772 
			&0.834 
			&0.789 
			&0.814 
			&0.822 
			&0.832 
			&0.800 
			&0.792 
			&0.806 
			&0.821 
			&0.827 
			& \color{reda} \textbf{0.859} 
			&0.824 
			&0.811 
			&0.816 
			&0.823 
			&0.800 
			&0.775
			&0.815 
			&0.821 
			& \color{myblue} \textbf{0.839 }
			& \color{mygreen} \textbf{0.843} 
			&0.827 
			
			\\
			& $S_m\uparrow$  
			& 0.837 
			&0.876 
			&0.845 
			&0.864 
			&0.862 
			&0.864 
			&0.853 
			&0.847 
			&0.854 
			&0.859 
			&0.870 
			& \color{reda} \textbf{0.883} 
			&0.856 
			&0.854 
			&0.858 
			&0.854 
			&0.833 
			&0.827
			&0.862 
			&0.868 
			& \color{mygreen} \textbf{0.881} 
			& \color{myblue} \textbf{0.877} 
			&0.871

			\\
			& $E_m\uparrow$    &
			0.876 
			& \color{myblue} \textbf{0.908} 
			&0.883 
			&0.901 
			&0.902 
			&0.906 
			&0.894 
			&0.883 
			&0.891 
			&0.899 
			&0.906 
			& \color{reda} \textbf{0.923} 
			&0.903 
			&0.892 
			&0.898 
			&0.897 
			&0.875 
			&0.869
			&0.901 
			&0.900 
			& \color{mygreen} \textbf{0.909} 
			& \color{mygreen} \textbf{0.909} 
			&0.897 
			
			\\
			&$\mathcal{M}\downarrow$  &   0.094 
			&0.066 
			&0.080 
			&0.072 
			&0.070 
			&0.066 
			&0.074 
			&0.085 
			&0.077 
			&0.073 
			&0.068 
			& \color{reda} \textbf{0.055} 
			&0.071 
			&0.079 
			&0.073 
			&0.071 
			&0.077 
			&0.099
			&0.071 
			&0.072 
			& \color{myblue} \textbf{0.064} 
			& \color{mygreen} \textbf{0.063} 
			&0.066 
         			\\
   				&$AUC\uparrow$ 
&.9614
&.9734
&.9468
&.9697
&.9668
&.9570
&.9624
&.9640
&.9727
&.9532
&\color{myblue} \textbf{.9741}
&.9611
&.9460
&.9662
&.9606
&.9576
&.9087
&.9475
&.9594
&\color{mygreen} \textbf{.9746}
&\color{reda} \textbf{.9754}
&.9681
&.9625
	
			\\

			\bottomrule[2pt]
		\end{tabular}
	}
	\setlength{\abovecaptionskip}{2pt}
	\label{tab:RGBDSOD_performance}
\end{table*}
\begin{table*}
\centering
	\scriptsize
	\setlength{\abovecaptionskip}{2pt}
	\caption{Efficiency comparisons of the top-performing methods in Tab.~\ref{tab:RGBDSOD_performance}. The best three and worst results are shown in {\color{reda}\textbf{red}}, {\color{mygreen}\textbf{green}}, {\color{myblue}\textbf{blue}} and  {\textbf{black}}, respectively.}

   \resizebox{\linewidth}{!}{
    \setlength\tabcolsep{2pt}
    \renewcommand\arraystretch{1}
    
 \begin{tabular}{r||cccccccccccccc}
\toprule[1.5pt]
		Model Nmae  &{CDNet$_{21}$} &{DFMNet$_{21}$} &{DSA2F$_{21}$} &{DCF$_{21}$}&{HAINet$_{21}$} &   {RD3D$_{21}$} &   {SPNet$_{21}$} & {A2DELE$_{20}$}&{DASNet$_{20}$}&{CoNet$_{20}$} & {MMFT} &{MMFT} & {MMFT} & {MMFT} \\
&~\cite{CDNet}&~\cite{DFMNet}&~\cite{DSA2F}&~\cite{DCF}&~\cite{HAINet}&~\cite{RD3D}&~\cite{SPNet}&~\cite{A2DELE}&~\cite{DASNet}&~\cite{CoNet}& \\
		Backbone  &VGG-16 &ResNet-34& VGG-19& ResNet-50& VGG-16& ResNet-50& Res2Net-50&  VGG-16& ResNet-50&ResNet-101& VGG-16& ResNet-50& Res2Net-50& ResNet-101 \\
	\midrule[1pt]
		Parameters (MB)$\downarrow$    & \color{myblue}\textbf{32.93} &\color{mygreen}\textbf{23.44} & 36.48 & {47.85} & 59.82 &{46.90}&\textbf{150.31}&\color{reda}\textbf{14.90}&{36.68}&43.66& {41.38}&53.12&{75.96}&72.13
		\\
		FLOPs (G)$\downarrow$     &{72.10} & 26.32 &{172.23} &\color{reda}\textbf{17.74} & \textbf{181.41} & 50.72 &{67.92}&39.07&{28.56}&{27.37}&72.82&\color{mygreen}\textbf{20.32}&25.89&\color{myblue}\textbf{25.50}
		\\
 \bottomrule[1.5pt]
	\end{tabular}

	}

	\setlength{\abovecaptionskip}{2pt}
	\label{tab:para_flops}

\end{table*}

\begin{table*}
	\setlength{\abovecaptionskip}{2pt}
	\caption{Quantitative evaluation on RGB SOD datasets. $\uparrow$ and $ \downarrow$ indicate that the larger scores and the smaller ones  are better, respectively. The best three scores are highlighted in {\color{reda} \textbf{red}}, {\color{mygreen} \textbf{green}} and {\color{myblue} \textbf{blue}}.
	}
	\renewcommand\tabcolsep{5pt} 
	\renewcommand\arraystretch{1}
	\label{tab:rgbsod_performance}
	\centering
	\resizebox{\textwidth}{!}  
	{
		\begin{tabular}{c|c|c|ccccc|ccccc|ccccc|ccccc|ccccc}
			\toprule[1.5pt]
			\multicolumn{1}{c|}{\multirow{2}*{Method}}&\multicolumn{1}{c|}{\multirow{2}*{Publication}}&\multicolumn{1}{c|}{\multirow{2}*{Backbone}} &\multicolumn{5}{c|}{\emph{DUTS}}&\multicolumn{5}{c|}{\emph{DUT-OMRON}}&\multicolumn{5}{c|}{\emph{ECSSD}}&\multicolumn{5}{c|}{\emph{HKU-IS}}&\multicolumn{5}{c}{\emph{PASCAL-S}}\\
			
			\cmidrule(r){4-8} \cmidrule(r){9-13} \cmidrule(r){14-18} \cmidrule(r){19-23} \cmidrule(r){24-28} 
			&& & $F_{\beta}^{max}\uparrow$  & $F_{\beta}^{w}\uparrow$ &  $S_m\uparrow $ &  $E_m\uparrow $ & $\mathcal{M}\downarrow$
			& $F_{\beta}^{max}\uparrow$  & $F_{\beta}^{w}\uparrow$ &  $S_m\uparrow $ &  $E_m\uparrow $ & $\mathcal{M}\downarrow$ 
			& $F_{\beta}^{max}\uparrow$  & $F_{\beta}^{w}\uparrow$ &  $S_m\uparrow $ &  $E_m\uparrow $ & $\mathcal{M}\downarrow$
			& $F_{\beta}^{max}\uparrow$  & $F_{\beta}^{w}\uparrow$ &  $S_m\uparrow $ &  $E_m\uparrow $ & $\mathcal{M}\downarrow$& $F_{\beta}^{max}\uparrow$  & $F_{\beta}^{w}\uparrow$ &  $S_m\uparrow $ &  $E_m\uparrow $ & $\mathcal{M}\downarrow$ \\
			\midrule[1pt]
			
			GateNet~\cite{GateNet}&ECCV 2020 & ResNet-50&0.888 	&0.809 	&0.884 &	0.903 &	0.040 &	0.818 &	0.729 	&0.837 &	0.868 &	0.055 &	0.945 &	0.894 &	0.920 	&0.943 	&0.040 &	0.933 &	0.881 &	0.915 &	0.954 	&0.033 &	0.881 &	0.804& 	0.854 &	0.882 &	0.071 
			\\
			ITSD~\cite{ITSD} &CVPR 2020& ResNet-50 &0.883 &	0.824 &	0.883 &	0.898 	&0.041 &	0.821 &	0.750 &	0.839 &	0.867 &	0.061 	&0.947 &	0.911 &	0.925 &	0.932 &	0.035 &	0.933 	&0.893 &	0.916 &	0.953 &	0.031 &	0.882 &	0.823 	&0.859 	&0.866 &	0.066 
			
			\\
			MINet~\cite{MINet} &CVPR 2020& ResNet-50&0.884 	&0.825& 	0.883 &	0.917 	&\color{myblue} \textbf{0.037} &	0.810 &	0.738 &	0.832 &	0.873 &	0.056 	&0.948 &	0.911 &	0.925 &	0.953 &	0.034 &	0.935 	&0.899 &	0.919 	&\color{mygreen} \textbf{0.961} &	\color{mygreen} \textbf{0.028} &	0.880 &	0.818 	&0.854 &	0.897& 	0.066 
			
			\\
			GCPA~\cite{GCPA} & AAAI 2020& ResNet-50&0.888 &	0.821 &	0.889 &	0.913 	&0.038 &	0.812 &	0.734 &	0.837 &	0.869 &	0.056 	&0.949 &	0.903 &	0.927 &	0.952 &	0.035& 	0.939 	&0.889 	&\color{myblue} \textbf{0.920} &	0.956 &	0.032 &	0.882 &	0.819 	&\color{mygreen} \textbf{0.864} &	0.899 &	0.064 
			\\
			F3Net~\cite{F3Net} &AAAI 2020 & ResNet-50&	0.891 &		0.835 &		0.887 &		0.918 	&	\color{mygreen} \textbf{0.035} &		0.813 &		0.747 &		0.837 &		0.876 &		0.053 	&	0.945 &		0.912 &		0.924 &		0.946 &		\color{myblue} \textbf{0.033} &		0.937 	&	0.900 &		0.916 &		0.959 	&	\color{mygreen} \textbf{0.028} &		0.882 &		0.823 	&	0.857 &		0.892 &		0.064 
			
			\\
			VST~\cite{VST}&ICCV 2021&T2T~\cite{T2T}&0.890 &	0.828 &	\color{reda} \textbf{0.895} &	0.916 	&\color{myblue} \textbf{0.037} &	\color{myblue} \textbf{0.825} &	0.755 &	\color{reda} \textbf{0.849} &	0.872 &	0.058 	&\color{myblue} \textbf{0.951} &	0.910 &	\color{mygreen} \textbf{0.932} &\color{mygreen} \textbf{0.957} &	\color{myblue} \textbf{0.033} &	\color{mygreen} \textbf{0.942} 	&0.898 &	\color{reda} \textbf{0.928} &	\color{mygreen} \textbf{0.961} &	\color{myblue} \textbf{0.029} 	&\color{mygreen} \textbf{0.890} &	0.827 	&\color{reda} \textbf{0.871} &\color{reda} \textbf{0.905} &	\color{myblue} \textbf{0.062} 
			
			\\
			SAMNet~\cite{SAMNet} &TIP 2021&SAM~\cite{SAMNet}&\color{myblue} \textbf{0.894} &	0.840 &	0.890 &	\color{myblue} \textbf{0.923} 	&\color{mygreen} \textbf{0.035} &	0.823 &	\color{myblue} \textbf{0.760} 	&\color{myblue} \textbf{0.844} &	\color{myblue} \textbf{0.885} &	\color{myblue} \textbf{0.051} 	&\color{mygreen} \textbf{0.952}& 	\color{mygreen} \textbf{0.920} &	\color{myblue} \textbf{0.928} &	\color{reda} \textbf{0.958} &	\color{mygreen} \textbf{0.032} &	\color{myblue} \textbf{0.940} 	&\color{reda} \textbf{0.905} &	\color{mygreen} \textbf{0.921} &	\color{reda} \textbf{0.963} &	\color{reda} \textbf{0.027} &	\color{myblue} \textbf{0.887} &	\color{myblue} \textbf{0.832} 	&\color{myblue} \textbf{0.862} &\color{myblue} \textbf{0.903} &	\color{myblue} \textbf{0.062} 
			
			\\
			Auto-MSFNet~\cite{Auto-MSFNet}&ACMMM 2021&ResNet-50 &0.877 &	\color{myblue} \textbf{0.841} &	0.876 	&\color{reda} \textbf{0.931} &	\color{reda} \textbf{0.034}& 	0.798 &	0.757 &	0.831 &	0.876 	&\color{mygreen} \textbf{0.050} &	0.941 &	\color{myblue} \textbf{0.916} &	0.914 &	\color{myblue} \textbf{0.954} &	\color{myblue} \textbf{0.033} 	&0.928 &	\color{mygreen} \textbf{0.904}& 	0.908 &	\color{myblue} \textbf{0.960} &	\color{reda} \textbf{0.027} &	0.877 	&\color{mygreen} \textbf{0.834} &	0.851 &	\color{reda} \textbf{0.905} &	\color{mygreen} \textbf{0.061} 
			
			\\
			SGL-KRN~\cite{SGL-KRN} &AAAI 2021&ResNet-50 &\color{mygreen} \textbf{0.898} 	&\color{reda} \textbf{0.847} 	&\color{myblue} \textbf{0.891} &	\color{mygreen} \textbf{0.926} 	&\color{reda} \textbf{0.034} &	\color{mygreen} \textbf{0.827}& 	\color{mygreen} \textbf{0.765} &	\color{mygreen} \textbf{0.845} &	\color{mygreen} \textbf{0.889} &\color{reda} \textbf{0.049} 	&0.946 &	0.910 &	0.923 &	0.942 &	0.036 &	0.937 	&\color{myblue} \textbf{0.902} &	0.918 &	\color{myblue} \textbf{0.960} &	\color{myblue} \textbf{0.029}& 	\color{myblue} \textbf{0.887} &	0.826 	&0.856 &	0.883 &	0.068
			\\
			\midrule[1pt]
			\rowcolor{mygray}
			\multicolumn{2}{c|}{\multirow{1}*{Ours}}&ResNet-50&\color{reda} \textbf{0.904} &	\color{mygreen} \textbf{0.844} 	&\color{mygreen} \textbf{0.893} 	&\color{reda} \textbf{0.931} &	\color{reda} \textbf{0.034} &	\color{reda} \textbf{0.831} &	\color{reda} \textbf{0.766} &	\color{reda} \textbf{0.849} 	&\color{reda} \textbf{0.893} &	\color{reda} \textbf{0.049} &	\color{reda} \textbf{0.960} 	&\color{reda} \textbf{0.925} 	&\color{reda} \textbf{0.936} &\color{reda} \textbf{0.958} 	&\color{reda} \textbf{0.029} &	\color{reda} \textbf{0.944}& 	0.901& 	\color{mygreen} \textbf{0.921} &	\color{mygreen} \textbf{0.961} 	&\color{myblue} \textbf{0.029} &	\color{reda} \textbf{0.894}& 	\color{reda} \textbf{0.837} &	\color{reda} \textbf{0.871}& 	\color{mygreen} \textbf{0.904} &\color{reda} \textbf{0.059} 
			
			\\
			\bottomrule[1.5pt]
		\end{tabular}
	}
\end{table*}
\begin{table}
	\setlength{\abovecaptionskip}{2pt}
	\caption{Quantitative comparison for salient contour extraction.
	}
	\label{tab:Salient_contour_performance}
	\centering
	
	\resizebox{\columnwidth}{!} 
	{
 
		\begin{tabular}{c|c|cc|cc|cc|cc|cc}
			\toprule[1.5pt]
			\multicolumn{1}{c|}{\multirow{2}*{Method}}&\multicolumn{1}{c|}{\multirow{2}*{Backbone}} &\multicolumn{2}{c|}{\emph{STERE}}&\multicolumn{2}{c|}{\emph{NJUD}}&\multicolumn{2}{c|}{\emph{NLPR}}&\multicolumn{2}{c|}{\emph{SIP}}&\multicolumn{2}{c}{\emph{LFSD}}\\
			
			\cmidrule(r){3-4} \cmidrule(r){5-6} \cmidrule(r){7-8} \cmidrule(r){9-10} \cmidrule(r){11-12}  
			& &
			$F_{\beta}^{max}\uparrow$   & $\mathcal{M}\downarrow$
			& $F_{\beta}^{max}\uparrow$   & $\mathcal{M}\downarrow$ 
			& $F_{\beta}^{max}\uparrow$   & $\mathcal{M}\downarrow$   
			& $F_{\beta}^{max}\uparrow$   & $\mathcal{M}\downarrow$  
			& $F_{\beta}^{max}\uparrow$   & $\mathcal{M}\downarrow$
			
			\\
			\midrule[1pt]
			RD3D~\cite{RD3D} & ResNet-50 & 0.446&	0.017&  0.480&	0.019&	0.435& {0.014}&  0.390&	0.017&	0.440&	0.019 \\
			SPNet~\cite{SPNet} &	Res2Net-50 & 0.393&	0.018 &  0.490&	0.018&	0.426&	0.014&  0.384&	0.018&	0.377&	0.022    \\
			\rowcolor{mygray}
			MMFT & ResNet-50&	\color{reda} \textbf{0.493}&	 {0.017}&   \color{reda} \textbf{0.501}&	 \color{reda} \textbf{0.017}&	 \color{reda} \textbf{0.452}& {0.013}& \color{reda} \textbf{0.407}&{0.016}&	 \color{reda} \textbf{0.473}&{0.024}\\
			\rowcolor{mygray}
			MMFT & Res2Net-50&{0.480}&	 \color{reda} \textbf{0.016}& {0.494}&	 \color{reda} \textbf{0.017}&{0.443}&	\color{reda} \textbf {0.012}& {0.394}&	 \color{reda} \textbf{0.015}&{0.459}&	 \color{reda} \textbf{0.017}\\
			
			\bottomrule[1.5pt]
		\end{tabular}
	}
\end{table}
\begin{table}
	\setlength{\abovecaptionskip}{2pt}
	\caption{Quantitative comparison for depth estimation.
	}
	\label{tab:depth_estimation_performance}
	\centering
	\resizebox{\columnwidth}{!}    
	{
\begin{tabular}{c|c|cccc|ccc}
			\toprule[1.5pt]
			\multicolumn{1}{c|}{\multirow{2}*{Method}} &\multicolumn{1}{c|}{\multirow{2}*{Backbone}} &\multicolumn{4}{c|}{{Error}} & \multicolumn{3}{c}{{Accuracy}}\\
			
			\cmidrule(r){3-6} \cmidrule(r){7-9} 
			&& RMSE $\downarrow$   & RMSE (log) $\downarrow$
			& Abs Rel $\downarrow$   & Sq ReL $\downarrow$ & P1 $\uparrow$   & P2 $\uparrow$   
			& P3 $\uparrow$   
			\\
			\midrule[1pt]
			DASNet~\cite{DASNet} & ResNet-50&0.378&0.053&	0.130&	0.078& 0.739&	0.920&	0.984 \\
			CoNet~\cite{CoNet} & ResNet-101&0.433&0.062&	0.149&	0.094& 0.677&	0.907&	0.984 \\
			TL-Depth~\cite{TL-Depth} & ResNet-101&0.387&0.056&	0.134&	0.081& 0.721&	0.911&	0.983 \\
			LAP-Depth~\cite{LAP-Depth} & ResNet-101&0.404&0.049&	 {0.155}&	{0.111}& \color{reda} \textbf{0.771}&  0.925&	0.982 \\
			\rowcolor{mygray}
			MMFT &ResNet-50& \color{reda} \textbf{0.353} & \color{reda} \textbf{0.049}&	\color{reda} \textbf {0.124}& \color{reda} \textbf{0.070}&   {0.749}&  \color{reda} \textbf{0.931}&	 \color{reda} \textbf{0.986}\\
			\rowcolor{mygray}
			MMFT &ResNet-101& {0.355} & {0.050}&	 \color{reda} \textbf{0.124}&{0.071}&   {0.750}&  {0.927}&	 {0.985}\\
			\bottomrule[1.5pt]
		\end{tabular}		
	}
	\vspace{-2.5mm}
\end{table}  
\begin{figure*}
	\includegraphics[width=\textwidth]{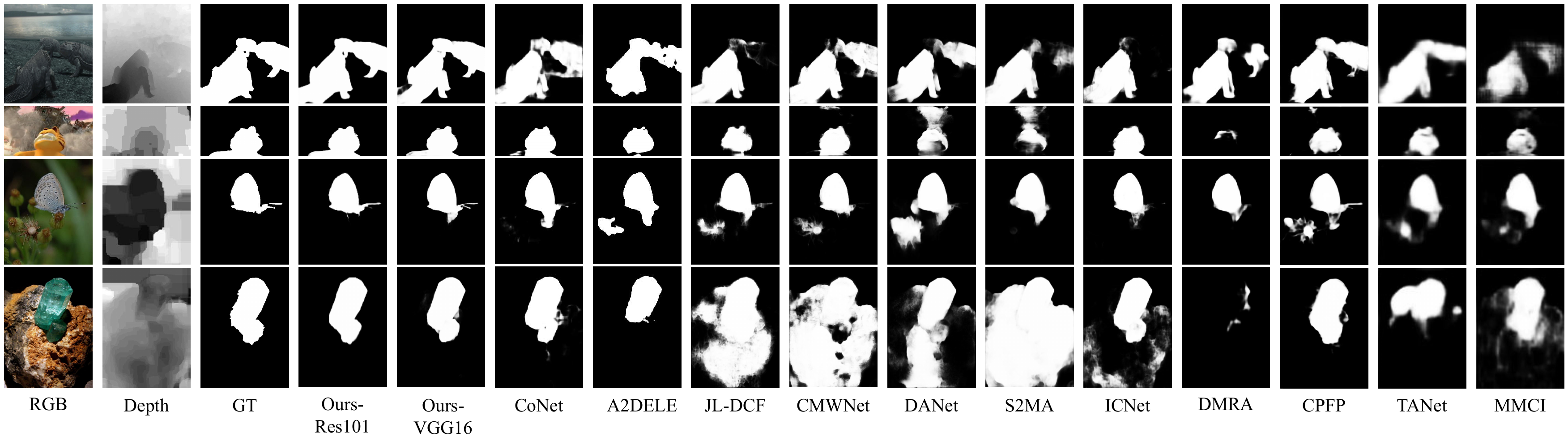}\\ 
	\centering
	\setlength{\abovecaptionskip}{-8pt}
	\caption{Visual comparison of different RGB-D SOD methods.} 
	\label{fig:rgbd_sod_comparison}
	\vspace{-3.5mm}
\end{figure*} 
\begin{figure*}
	\centering
	\includegraphics[width=0.8\linewidth]{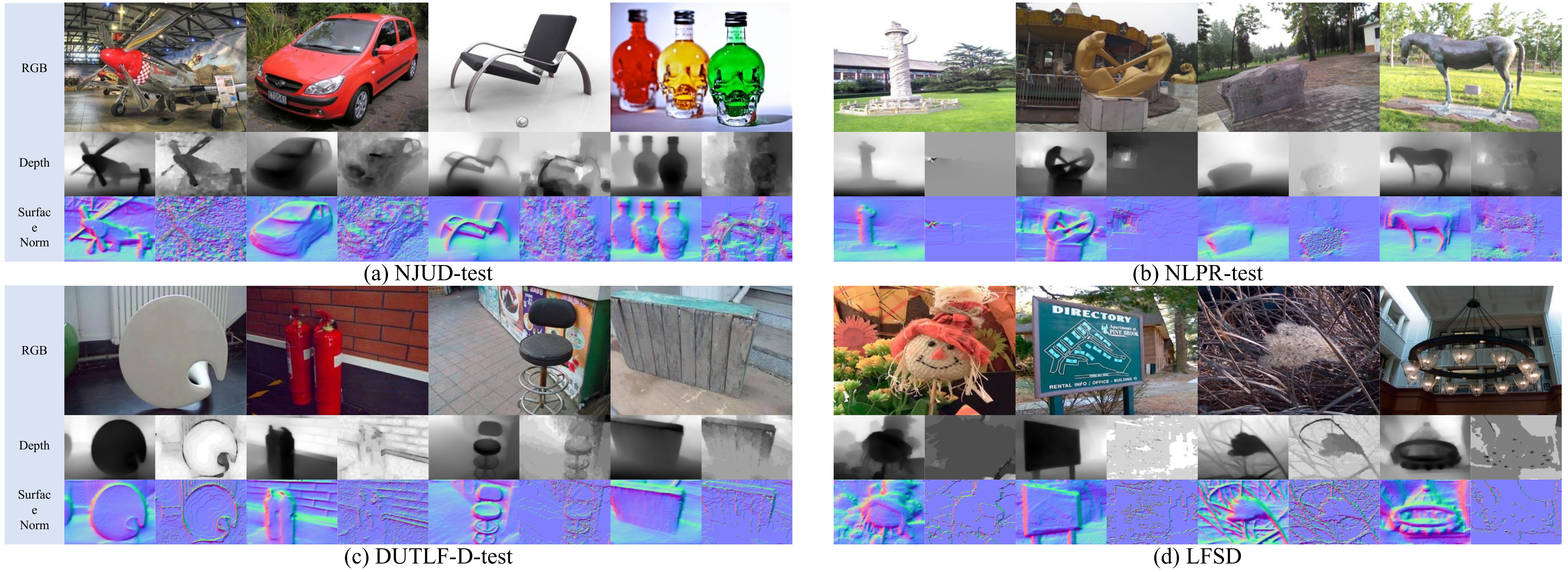}\\ 
	\centering
	\caption{Visual comparison of the predicted depth maps (Left) and the original ones (Right) provided by the RGB-D SOD dataset.} 
	\label{fig:njud_nlpr_dut_lfsd_depth_surface}
	\vspace{-5.5mm}
\end{figure*}  
\begin{figure*}
	\centering
	\includegraphics[width=0.8\textwidth]{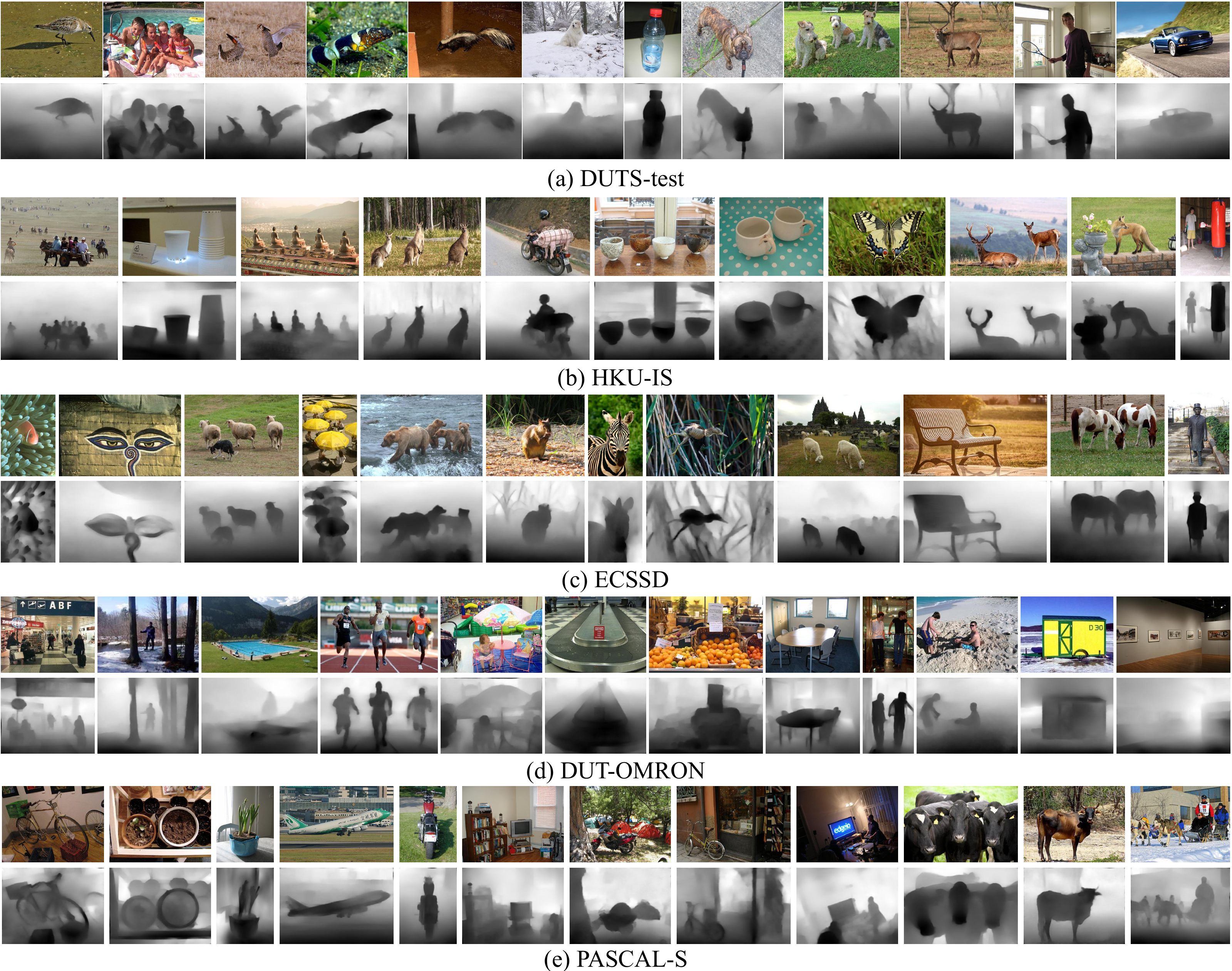}
	\centering
	\caption{The predicted depth maps on RGB SOD datasets~\cite{DUTS,DUT-OMRON,HKU-IS,ECSSD,PASCAL-S}.} 
	\label{fig:rgbsod_depth}
\end{figure*}  
\begin{table*}
	\setlength{\abovecaptionskip}{2pt}
	\caption{Qualitative evaluation of applying the predicted depth map to existing depth-based RGB-D SOD methods. ``$\ast$'' and ``$\ddagger$'' mean using the original depth map and ours, respectively.
	}
	\label{tab:applying_depth_to_rgbdsod}
	\centering
	\resizebox{\textwidth}{!}  
	{	\begin{tabular}{c|ccccc|ccccc|ccccc|ccccc|ccccc}
			\toprule[1.5pt]
			\multicolumn{1}{c|}{\multirow{2}*{Method}} &\multicolumn{5}{c|}{\emph{SIP}}&\multicolumn{5}{c|}{\emph{STERE}}&\multicolumn{5}{c|}{\emph{NJUD}}&\multicolumn{5}{c|}{\emph{NLPR}}&\multicolumn{5}{c}{\emph{LFSD}}\\
			
			\cmidrule(r){2-6} \cmidrule(r){7-11} \cmidrule(r){12-16} \cmidrule(r){17-21} \cmidrule(r){22-26} 
			& $F_{\beta}^{max}\uparrow$  & $F_{\beta}^{w}\uparrow$ &  $S_m\uparrow $ &  $E_m\uparrow $ & $\mathcal{M}\downarrow$
			& $F_{\beta}^{max}\uparrow$  & $F_{\beta}^{w}\uparrow$ &  $S_m\uparrow $ &  $E_m\uparrow $ & $\mathcal{M}\downarrow$ 
			& $F_{\beta}^{max}\uparrow$  & $F_{\beta}^{w}\uparrow$ &  $S_m\uparrow $ &  $E_m\uparrow $ & $\mathcal{M}\downarrow$
			& $F_{\beta}^{max}\uparrow$  & $F_{\beta}^{w}\uparrow$ &  $S_m\uparrow $ &  $E_m\uparrow $ & $\mathcal{M}\downarrow$ 
			& $F_{\beta}^{max}\uparrow$  & $F_{\beta}^{w}\uparrow$ &  $S_m\uparrow $ &  $E_m\uparrow $ & $\mathcal{M}\downarrow$ 
			\\
			\midrule[1pt]

			Two-stream FPN$\ast$ & 0.886&	0.810&	0.847&	0.890&	0.061&	0.882&	0.815&	0.865& {0.899}&	0.051& 0.893&	0.834&	0.868&	0.901&	0.052&	0.896&	0.845&	0.889&	0.920&	0.035 &0.844 &	0.767 &	0.827 &	0.864 &	0.098 \\
			\rowcolor{mygray}
			Two-stream FPN$\ddagger$& \color{reda} \textbf{0.903}&	\color{reda} \textbf{0.840}&	\color{reda} \textbf{0.881}&	\color{reda} \textbf{0.919}&	\color{reda} \textbf{0.045}&	\color{reda} \textbf{0.911}&	\color{reda} \textbf{0.850}&	\color{reda} \textbf{0.890}&	\color{reda} \textbf{0.914}&	\color{reda} \textbf{0.042}& \color{reda} \textbf{0.909}&	\color{reda} \textbf{0.857}&	\color{reda} \textbf{0.894}&	\color{reda} \textbf{0.921}&	\color{reda} \textbf{0.043}&	\color{reda} \textbf{0.914}&	\color{reda} \textbf{0.859}&	\color{reda} \textbf{0.918}&	\color{reda} \textbf{0.945}&	\color{reda} \textbf{0.029} 
			&\color{reda} \textbf{0.855} &	\color{reda} \textbf{0.787} &	\color{reda} \textbf{0.839} &	\color{reda} \textbf{0.880} &	\color{reda} \textbf{0.088} 
			\\
			RD3D$\ast$~\cite{RD3D} & 0.906 &	0.845 &	0.885 &	0.924 &	0.048&  0.917 &	0.871& 	0.911 &	0.944& 	0.037 
			&0.923 	&0.886 	&0.916 	&0.942 &	0.037 &0.927 	&0.889 &	0.929 &	\color{reda} \textbf{0.959} &	0.022 &0.879 &	0.816 &	0.858 &	0.898 &	0.073 
			\\
			\rowcolor{mygray}
			RD3D$\ddagger$~\cite{RD3D} & \color{reda} \textbf{0.925} &	\color{reda} \textbf{0.879} &	\color{reda} \textbf{0.905} &	\color{reda} \textbf{0.940} &	\color{reda} \textbf{0.038} &	\color{reda} \textbf{0.932} &	\color{reda} \textbf{0.893} &	\color{reda} \textbf{0.922} &	\color{reda} \textbf{0.948} &	\color{reda} \textbf{0.030} &\color{reda} \textbf{0.937} &	\color{reda} \textbf{0.902} &	\color{reda} \textbf{0.924} &	\color{reda} \textbf{0.947} &	\color{reda} \textbf{0.031} &	\color{reda} \textbf{0.928} &	\color{reda} \textbf{0.893} &	\color{reda} \textbf{0.929} &	{0.957} & \color{reda} \textbf{0.022}&	\color{reda} \textbf{0.888} &	\color{reda} \textbf{0.828} &	\color{reda} \textbf{0.867} &	\color{reda} \textbf{0.898} & \color{reda} \textbf{0.068}\\
			SPNet$\ast$~\cite{SPNet} &0.916 &	0.868 &	0.894 	&0.931 	&0.043 &0.915 &	0.873 &	0.907 &	0.942 	&0.037 &0.935 &	0.906 	&0.924 	&0.953 &	0.029  &0.926 &	0.896 &	0.927 	&0.959 &	0.021 &0.881 &	0.823 &	0.854 &	0.897 	&0.071 
			
			\\
			\rowcolor{mygray}
			SPNet$\ddagger$~\cite{SPNet}& \color{reda} \textbf{0.928}&	\color{reda} \textbf{0.883}&	\color{reda} \textbf{0.907}&	\color{reda} \textbf{0.942}&	\color{reda} \textbf{0.037}&	\color{reda} \textbf{0.931}&	\color{reda} \textbf{0.894}&	\color{reda} \textbf{0.922}&	\color{reda} \textbf{0.952}&	\color{reda} \textbf{0.030}& \color{reda} \textbf{0.941}&	\color{reda} \textbf{0.910}&	\color{reda} \textbf{0.928}&	\color{reda} \textbf{0.954}&	\color{reda} \textbf{0.029}&	\color{reda} \textbf{0.931}&	\color{reda} \textbf{0.898}&	\color{reda} \textbf{0.930}&	\color{reda} \textbf{0.961}&	\color{reda} \textbf{0.021} &	\color{reda} \textbf{0.882}&	\color{reda} \textbf{0.826}&	\color{reda} \textbf{0.866}&	\color{reda} \textbf{0.901}&	\color{reda} \textbf{0.065}\\
			\bottomrule[1.5pt]
		\end{tabular}
	}
\end{table*}
\begin{figure}
	\includegraphics[width=\linewidth]{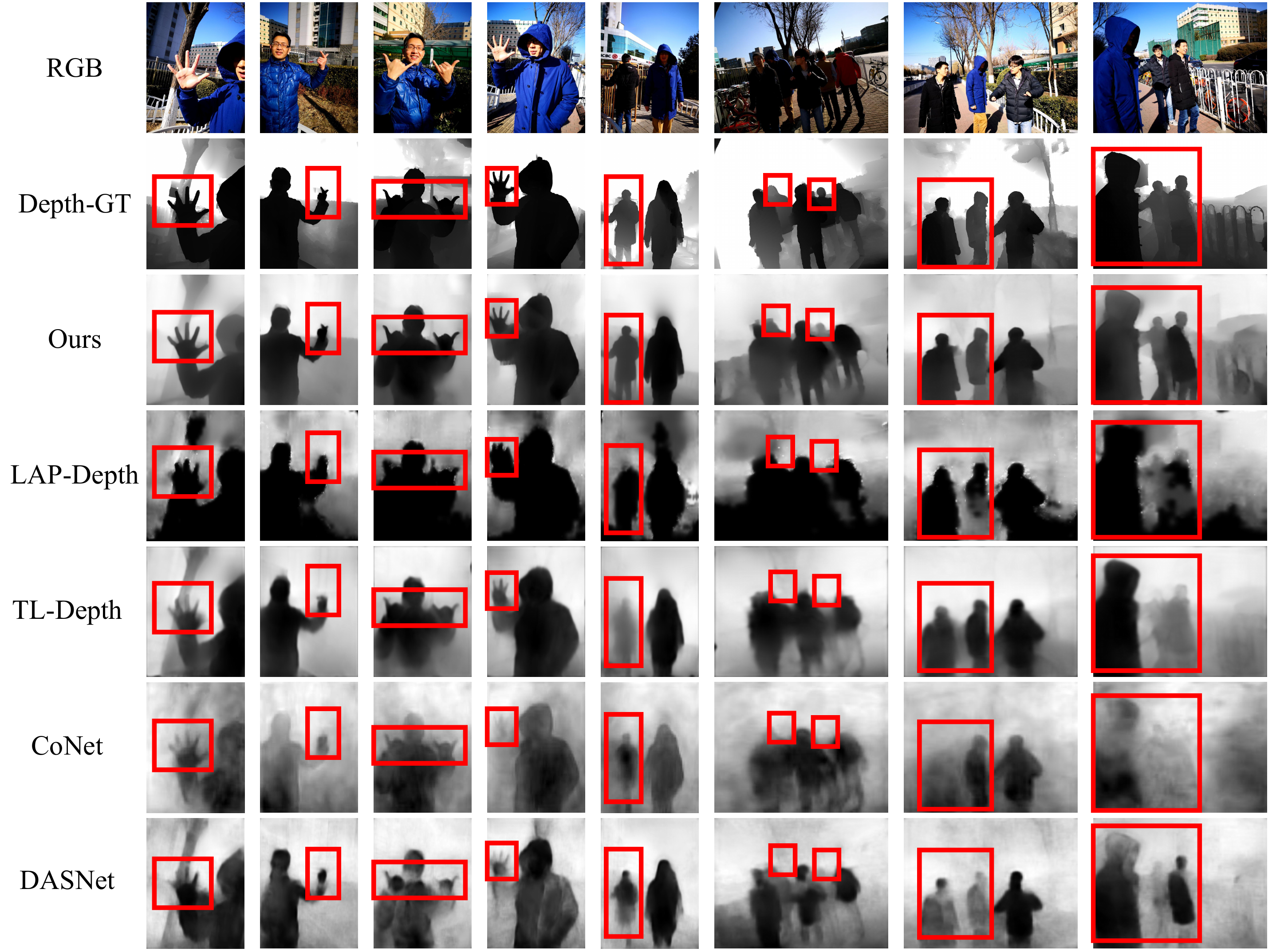}
	\setlength{\abovecaptionskip}{-2pt}
	\centering
	\caption{Visual comparisons with several depth estimation methods~\cite{TL-Depth,LAP-Depth} and CoNet~\cite{CoNet} on the SIP dataset~\cite{SIP}.} 
	\label{fig:SIP_depth_comparison}
	\vspace{-2.5mm}
\end{figure}
\noindent\textbf{\textit{Depth Evaluation.}} Tab.~\ref{tab:depth_estimation_performance} shows the results of depth estimation on the SIP dataset in terms of seven common metrics. We compare the proposed method with two depth estimation methods TL-Depth~\cite{TL-Depth} and LAP-Depth~\cite{LAP-Depth} and two depth-free RGB-D SOD methods CoNet~\cite{CoNet} and DASNet~\cite{DASNet}. It can be seen that the proposed method outperforms these models across six metrics. In addition, in order to show the benefits of the predicted depth maps, we apply them to existing depth-based RGB-D SOD methods instead of using the original ones and retrain them, which include the baseline (two-stream FPN) used in~\cite{HDFNet,S2MA,JLDCF,RD3D,SPNet} and existing top 
two methods SPNet~\cite{SPNet} and RD3D~\cite{RD3D}. As shown in Tab.~\ref{tab:applying_depth_to_rgbdsod}, all these models obtain a considerable performance gain.

\subsubsection{Qualitative Evaluation}
~\\
\textbf{\textit{Saliency Evaluation.}}
Fig.~\ref{fig:rgbd_sod_comparison} illustrates several visual results of different RGB-D SOD approaches. The proposed method yields the results closer to the ground truth in various challenging scenarios. In the $1^{st}$ and $2^{nd}$ rows, we show two examples which have fuzzy depth maps. From the results, it can be observed that our method can segment the objects well while the other methods more or less lose similar areas inside or around salient objects.  In the $3^{rd}$ and $4^{th}$ rows, 
it can be seen that our method has more obvious advantages when handling internal hollow areas, and our saliency maps have sharper contours with the help of the contour decoder.
\\ 
\textbf{\textit{Depth Evaluation.}} 
The proposed method can predict the denoised depth map. As shown in the Fig.~\ref{fig:STERE_depth_surfacenorm} and Fig.~\ref{fig:njud_nlpr_dut_lfsd_depth_surface}, compared with the original depth map provided by  RGB-D SOD dataset, the advantage of our predicted depth map is very significant. To intuitively show the depth map, we utilize it to generate a surface norm map that measures the smoothness of the surface of an object. We can see that the low-quality depth map contains serious Gaussian-like noise and checkerboard pattern. In contrast, the key appearance information such as the contour of the object is better embodied in our depth map. We further qualitative evaluate our depth results on the high-quality SIP~\cite{SIP} dataset, as shown in Fig.~\ref{fig:SIP_depth_comparison}. It can be seen that our predicted depth maps are closer to the ground-truth and retain the human contours very well (See the $1^{st}$ - $4^{th}$ columns). For complex scenes with multiple objects, our method is still able to  perceive the positions of all objects while accurately reflecting the relative depth relationships among them (See the $5^{th}$ - $8^{th}$ columns). Besides, we also show some depth results on several RGB SOD datasets, as shown in Fig.~\ref{fig:rgbsod_depth}. It can be seen that both multiple object and single object cases are accurately measured and even overcome the influence of shadows, which indicates good generalization ability. 
\begin{figure}
	\includegraphics[width=\linewidth]{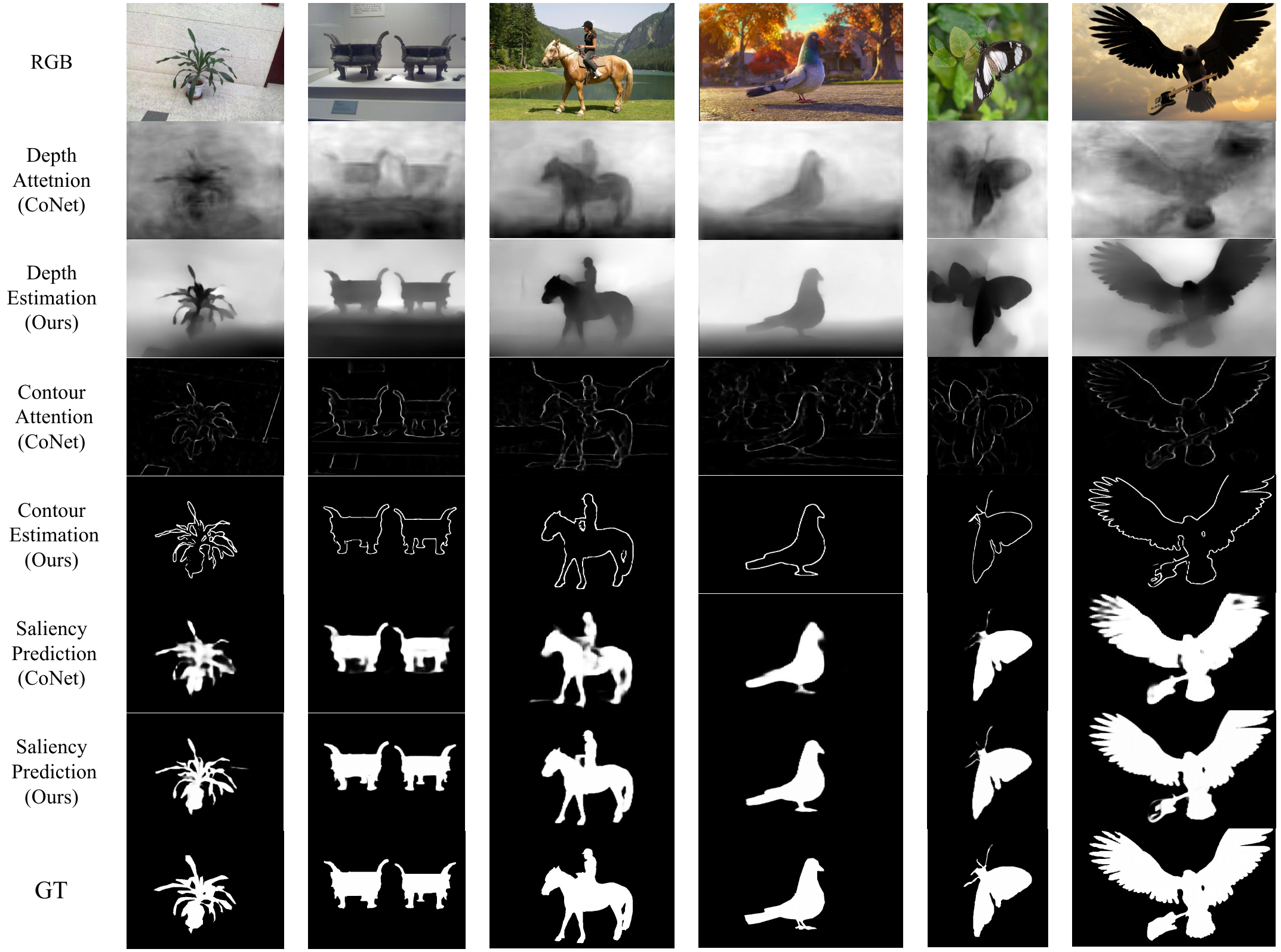}
	\setlength{\abovecaptionskip}{-10pt}
	\centering
	\caption{Visual comparisons between ours and CoNet~\cite{CoNet} for depth estimation,  contour  and saliency prediction.} 
	\label{fig:comparison_conet}
	\vspace{-5.5mm}
\end{figure}
\begin{table*}
	\setlength{\abovecaptionskip}{2pt}
	\caption{Ablation  experiments  of  each component for salient object detection. Model 1: the FPN Baseline (Saliency Decoder).  Model 2: + Depth Estimation Decoder. Model 3: + Contour Estimation Decoder. Model 4: Without Squeeze-and-Expand in decoder block. Model 5 - Model 9: + Multi-modal Transformer (1, 2, 4, 6 and 8 iterations). Model 10: + Modality-specific Filter. Model 11: Replace the MFT with general Non-local+Convs. Model 12: Replace the MFT with Convs.
	}
	
	\label{tab:ablation_study_rgbdsod}
	\centering
	\resizebox{\textwidth}{!}  
	{\begin{tabular}{c|c|c|ccccc|ccccc|ccccc|ccccc|ccccc}
			\toprule[1.5pt]
			\multicolumn{1}{c|}{\multirow{2}*{Method}}&\multicolumn{2}{c|}{\emph{}} &\multicolumn{5}{c|}{\emph{SIP}}&\multicolumn{5}{c|}{\emph{STERE}}&\multicolumn{5}{c|}{\emph{NJUD}}&\multicolumn{5}{c|}{\emph{NLPR}}&\multicolumn{5}{c}{\emph{LFSD}}\\
			
			\cmidrule(r){2-3} \cmidrule(r){4-8} \cmidrule(r){9-13} \cmidrule(r){14-18} \cmidrule(r){19-23} \cmidrule(r){24-28} 
			& Params (MB)$\downarrow$ & FLOPs (GB)$\downarrow$ & $F_{\beta}^{max}\uparrow$  & $F_{\beta}^{w}\uparrow$ &  $S_m\uparrow $ &  $E_m\uparrow $ & $\mathcal{M}\downarrow$
			& $F_{\beta}^{max}\uparrow$  & $F_{\beta}^{w}\uparrow$ &  $S_m\uparrow $ &  $E_m\uparrow $ & $\mathcal{M}\downarrow$ 
			& $F_{\beta}^{max}\uparrow$  & $F_{\beta}^{w}\uparrow$ &  $S_m\uparrow $ &  $E_m\uparrow $ & $\mathcal{M}\downarrow$
			& $F_{\beta}^{max}\uparrow$  & $F_{\beta}^{w}\uparrow$ &  $S_m\uparrow $ &  $E_m\uparrow $ & $\mathcal{M}\downarrow$& $F_{\beta}^{max}\uparrow$  & $F_{\beta}^{w}\uparrow$ &  $S_m\uparrow $ &  $E_m\uparrow $ & $\mathcal{M}\downarrow$ \\
			\midrule[1pt]

			Model 1 & 53.54& 	17.98&  0.860& 	0.766& 	0.831 &	0.833 &	0.062 &	0.868 &	0.784 &	0.849 &	0.862 &	0.051 &	0.872&	0.799&	0.858&	0.847&	0.056&	0.873 &	0.810 &	0.874 &	0.882 	&0.038 &	0.800& 	0.713& 	0.777 &	0.801 &	0.117 
			\\
			Model 2 & 55.54&	21.35& 0.884  & 	0.801  & 	0.857 &  	0.863  & 	0.054  & 	0.890  & 	0.817 &  	0.873  & 	0.881  & 	0.044  & 	0.888 & 	0.834 & 	0.878 & 	0.872 & 	0.047	 & 0.890  & 	0.827  & 	0.885  & 	0.899  & 	0.034  & 	0.827  & 	0.741  & 	0.804  & 	0.829  & 	0.105 
			\\
			\rowcolor{mygray}
			Model 3 & 57.54& 24.71& \color{myblue} \textbf{0.896}&	\color{myblue} \textbf{0.812}&	\color{myblue} \textbf{0.866}&	\color{myblue} \textbf{0.874}&	\color{myblue} \textbf{0.051}&	\color{myblue} \textbf{0.899}&	\color{myblue} \textbf{0.830}&	\color{myblue} \textbf{0.887}&	\color{myblue} \textbf{0.895}&	\color{myblue} \textbf{0.042}& \color{myblue} \textbf{0.893}&	\color{myblue} \textbf{0.846}&	\color{myblue} \textbf{0.888}&	\color{myblue} \textbf{0.892}&	\color{myblue} \textbf{0.044}&	\color{myblue} \textbf{0.896}&	\color{myblue} \textbf{0.840}&	\color{myblue} \textbf{0.894}&	\color{myblue} \textbf{0.910}&	\color{myblue} \textbf{0.032}&\color{myblue} \textbf{0.840} &	\color{myblue} \textbf{0.762} &	\color{myblue} \textbf{0.819} &	\color{myblue} \textbf{0.840} &	\color{myblue} \textbf{0.099} 
			\\
			Model 4 & 56.62& 23.28& 0.887 &	0.787 &	0.853 &	0.854 &	0.056&	0.887 &	0.814 &	0.877 &	0.883&	0.044 &	0.884&	0.832&	0.880&	0.878&	0.046&	0.890 &	0.823 	&0.883 &	0.896 	&0.035 &	0.818&	0.726&	0.796&	0.831	&0.108
			
			\\
			
			\toprule[1pt]
			\toprule[1pt]
			Model 5 & 60.30&	24.89& 0.910 &	0.841& 	0.882 &	0.901 &	0.046 &	0.911 &	0.855 &	0.900 &	0.916 &	0.038 &	0.910 &	0.869& 	0.902& 	0.911 &	0.038 	&0.909 &	0.860 &	0.911 &	0.932 &	0.028 &	0.851 &	0.792& 	0.834& 	0.855 &	0.083 
			\\
			Model 6 & 63.94&	25.12& 0.913 &	0.850 &	0.886 &	0.909 &	0.044 &	0.917 &	0.866 &	0.907 &	0.928 &	0.037 &	0.918 &	0.882 &	0.908& 	0.923 &	0.036 	&0.912 &	0.867 &	0.915 &	0.940 &	0.027 &	0.856 &	0.799 &	0.838& 	0.862 &	0.080 
			\\
			Model 7 & 71.22&	25.58&0.918 &	0.857 &	0.893 &	0.917 &	0.043 &	0.921 &	0.872 &	0.908 &	0.936 &	0.036 &	0.924 &	0.882 &	0.915 &	0.930& 	0.036 	&0.918 &	0.873 &	0.920 &	0.944 &	0.026 &	0.864 &	0.808 &	0.846 &	0.874 &	0.075 
			\\
			Model 8 & 78.51&	26.05& 0.920 &	0.863 &	0.897 	&0.920 &	0.042 &	0.924 &	0.879& 	0.912 &	0.943 &	0.035 &	0.928 &	0.886 &	0.917 &	0.935 &	0.034 	&0.921 &	0.879 	&0.922 	&0.950 &	0.024& 	0.872 &	0.811 &	0.852& 	0.883 &	0.072 
			\\
			Model 9 & 85.79&	26.52& 0.917 &	0.863 &	0.895 &	0.919 &	0.042 &	0.924 &	0.878 &	0.912 &	0.942 &	0.035 &	0.929 &	0.884 &	0.913 &	0.934 &	0.034 	&0.921 &	0.880 	&0.920 &	0.949& 	0.024& 	0.872 &	0.808 &	0.854 &	0.884 &	0.072 
			\\
			\rowcolor{mygray}
			Model 10  & 75.96&	25.89& \color{reda} \textbf{0.931}&	\color{reda} \textbf{0.881}&	\color{reda} \textbf{0.909}&	\color{reda} \textbf{0.943}&	\color{reda} \textbf{0.038}&	\color{reda} \textbf{0.935}&	\color{reda} \textbf{0.894}&	\color{reda} \textbf{0.925}&	\color{reda} \textbf{0.953}&	\color{reda} \textbf{0.031} &\color{reda} \textbf{0.940}	&\color{reda} \textbf{0.905}&	\color{reda} \textbf{0.927}&	\color{reda} \textbf{0.950}&	\color{reda} \textbf{0.031}&	\color{reda} \textbf{0.931}&	\color{reda} \textbf{0.893}&	\color{reda} \textbf{0.933}&	\color{reda} \textbf{0.958}&	\color{reda} \textbf{0.022} & \color{reda} \textbf{0.897}& 	\color{reda} \textbf{0.843} &	\color{reda} \textbf{0.877} &	\color{reda} \textbf{0.909} &	\color{reda} \textbf{0.063} 
			\\
			Model 11 & 76.20&	26.12&  0.908 &	0.852 &	0.883 &	0.907 &	0.045 &	0.921 &	0.870 &	0.913 &	0.932 &	0.037 	&0.917& 	0.877 &	0.916 &	0.921 &	0.037 	&0.911 &	0.870 &	0.913 &	0.937& 	0.028 &	0.855 &	0.803 &	0.835 	&0.857 	&0.080 
			\\
			Model 12 & 76.40&	27.45&  0.894 &	0.808 &	0.868 &	0.879 &	0.052&	0.901 &	0.834 &	0.893 &	0.899 &	0.042 &	0.890 &	0.844 &	0.889 &	0.891 &	0.045 	&0.898 &	0.848& 	0.896 &	0.914 &	0.031 &	0.833 &	0.754 &	0.813& 	0.840 &	0.102 
			\\
			\bottomrule[1.5pt]
		\end{tabular}
	}
	   \vspace{-3.5mm}
\end{table*}
\begin{table}
	\setlength{\abovecaptionskip}{2pt}
	\caption{Ablation  experiments  of  each component for depth estimation. Model 1: the FPN Baseline (Depth Estimation Decoder).  Model 2: + Saliency Decoder. Model 3: + Contour Estimation Decoder. The others ``Model *'' correspond to the same model in Tab.~\ref{tab:ablation_study_rgbdsod}.
	}
	\label{tab:ablation_study_depth_estimation}
	\centering
	\resizebox{\columnwidth}{!}  
	{
		\begin{tabular}{c|cccc|ccc}
			\toprule[1.5pt]
			\multicolumn{1}{c|}{\multirow{2}*{Method}} &\multicolumn{4}{c|}{{Error}} & \multicolumn{3}{c}{{Accuracy}}\\
			
			\cmidrule(r){2-5} \cmidrule(r){6-8} 
			& RMSE $\downarrow$   & RMSE (log) $\downarrow$
			& Abs Rel $\downarrow$   & Sq ReL $\downarrow$ & P1 $\uparrow$   & P2 $\uparrow$   
			& P3 $\uparrow$   
			\\
			\midrule[1pt]
			Model 1&0.424&	0.085&	0.180&	0.120&  0.670&	0.851&	0.970  \\
			Model 2&0.394&	0.075&	0.166&	0.096&  0.710&	0.882&	0.974 \\
			\rowcolor{mygray}
			Model 3&\color{myblue} \textbf{0.389}&	\color{myblue} \textbf{0.070}&\color{myblue} \textbf{0.161}&\color{myblue}\textbf{0.093}& \color{myblue} \textbf{0.718}&	\color{myblue} \textbf{0.890}&	\color{myblue} \textbf{0.976} \\
			Model 4 &0.400&	0.078&	0.171&	0.102&  0.700&	0.871&	0.972 \\
			\hline
			\hline
			Model 5 &0.368&0.060&0.147&0.082& 0.739&  0.904 & 0.980\\
			Model 6 &0.362&0.056&0.143&0.078& 0.745&  0.912 & 0.980\\
			Model 7 &0.358&0.054&0.140&0.076& 0.748&  0.917 & 0.982\\
			Model 8 &0.355&0.052&0.137&0.075& 0.752&  0.922 & 0.982\\
			Model 9 &0.356&0.052&0.138&0.074& 0.750&  0.921 & 0.981\\
			\rowcolor{mygray}
			Model 10 & \color{reda} \textbf{0.343}& \color{reda} \textbf{0.048}& \color{reda} \textbf{0.120}& \color{reda} \textbf{0.068}& \color{reda} \textbf{0.763}& \color{reda} \textbf{0.931}  & \color{reda} \textbf{0.984}\\
			Model 11 &0.372&0.064&0.150&0.085& 0.734&  0.900 & 0.978\\
			Model 12 &0.387&0.072&	0.170&	0.105&  0.720&	0.874&	0.973 \\
			\bottomrule[1.5pt]
		\end{tabular}
	}
	\vspace{-2.5mm}
\end{table}

\begin{table}
	\setlength{\abovecaptionskip}{2pt}
	\caption{Ablation  experiments  of  each component for contour estimation. Model 1: the FPN Baseline (Contour Estimation Decoder).  Model 2: + Saliency Decoder. Model 3: + Depth Estimation Decoder. The others ``Model *'' correspond to the same model in Tab.~\ref{tab:ablation_study_rgbdsod}.
	}
	\label{tab:ablation_study_salient_contour}
	\centering
	\resizebox{\columnwidth}{!}  
	{
		\begin{tabular}{c|cc|cc|cc|cc|cc}
			\toprule[1.5pt]
			\multicolumn{1}{c|}{\multirow{2}*{Method}} &\multicolumn{2}{c|}{\emph{STERE}}&\multicolumn{2}{c|}{\emph{NJUD}}&\multicolumn{2}{c|}{\emph{NLPR}}&\multicolumn{2}{c|}{\emph{SIP}}&\multicolumn{2}{c}{\emph{LFSD}}\\
			
			\cmidrule(r){2-3} \cmidrule(r){4-5} \cmidrule(r){6-7} \cmidrule(r){8-9} \cmidrule(r){10-11}  
			&
			$F_{\beta}^{max}\uparrow$   & $\mathcal{M}\downarrow$
			& $F_{\beta}^{max}\uparrow$   & $\mathcal{M}\downarrow$ 
			& $F_{\beta}^{max}\uparrow$   & $\mathcal{M}\downarrow$   
			& $F_{\beta}^{max}\uparrow$   & $\mathcal{M}\downarrow$  
			& $F_{\beta}^{max}\uparrow$   & $\mathcal{M}\downarrow$
			
			\\
			\midrule[1pt]
			Model 1  &0.414&0.021&0.432&0.019&0.381&0.019&0.332&0.019&0.393&0.026\\
			Model 2 &0.436&0.018&0.450&0.021&0.400&0.017&0.348&0.017&0.410&0.023\\
			\rowcolor{mygray}
			Model 3 & \color{myblue} \textbf{0.442}&\color{myblue} \textbf{0.018}&\color{myblue} \textbf{0.459}&\color{myblue} \textbf{0.020}&\color{myblue} \textbf{0.406}&\color{myblue} \textbf{0.016}&\color{myblue} \textbf{0.356}&\color{myblue} \textbf{0.016}&\color{myblue} \textbf{0.417}&\color{myblue} \textbf{0.022}\\
			Model 4  &0.430&0.019&0.448&0.022&0.394&0.018&0.340&0.018&0.405&0.023
			\\
			\hline
			\hline
			Model 5&0.454&0.016&0.472&0.018&0.420&0.015&0.366&0.017&0.437&0.020\\
			Model 6&0.460&0.017&0.477&0.019&0.422&0.015&0.372&0.016&0.440&0.019\\
			Model 7&0.465&0.018&0.479&0.019&0.427&0.014&0.380&0.017&0.442&0.019\\
			Model 8&0.468&0.018&0.484&0.019&0.430&0.014&0.385&0.017&0.445&0.019
			\\
			Model 9 &0.466&0.018&0.484&0.019&0.429&0.014&0.386&0.017&0.444&0.019
			\\
			\rowcolor{mygray}
			Model 10 &	 \color{reda} \textbf{0.480}&	 \color{reda} \textbf{0.016}& 	 \color{reda} \textbf{0.494}&	 \color{reda} \textbf{0.017}&	 \color{reda} \textbf{0.443}&	\color{reda} \textbf {0.012}&	 \color{reda} \textbf {0.394}&	 \color{reda} \textbf{0.015}&	 \color{reda} \textbf{0.459}&	 \color{reda} \textbf{0.017}\\
			Model 11&0.459&0.017&0.470&0.019&0.418&0.018&0.364&0.017&0.430&0.020\\
			Model 12 &0.440&0.018&0.457&0.020&0.406&0.017&0.350&0.017&0.417&0.022\\
			
			\bottomrule[1.5pt]
		\end{tabular}
	}
	\vspace{-5.5mm}
\end{table}
\subsection{Comprehensive Comparison with CoNet} \label{sec:Comprehensive_Comparison_with CoNet}
CoNet~\cite{CoNet} unidirectionally utilizes depth attention and contour attention to help RGB-D salient object detection, which is not an implementation of multi-task learning. In order to fully demonstrate our superiority and  difference, we visualize some results in Fig.~\ref{fig:comparison_conet}. Benefiting from the multi-task design and the \textit{Squeeze-and-Expand} strategy, we obtain better predictions on the three tasks. The smooth body, sharp contour and accurate localization are fully reflected in each task.
\subsection{Ablation Studies for Salient Object Detection}\label{sec:AbStd}
In this section, we show the effectiveness of each component for the salient object detection. All the experiments are based on the Res2Net-50 backbone. 
\\
\textbf{Multi-task Learning.} 
We take the FPN with the RGB input and a saliency decoder branch as the baseline. The depth estimation and contour extraction branches are then added to the baseline, respectively. We apply the \textit{Squeeze-and-Expand} (SE) structure in each decoder branch. As shown in  Tab.~\ref{tab:ablation_study_rgbdsod}, Model $2$ vs. Model $1$ and Model $3$ vs. Model $2$ show the effectiveness of  depth estimation and salient contour extraction for saliency prediction, respectively.
It can be seen that the multi-task network (Model $3$) achieves a significant improvement compared to the baseline with the  gain of $5.87\%$ and $17.65\%$ in terms of $F_{\beta}^{w}$ and MAE on the challenging  STERE dataset. In addition, we remove all SEs to obtain Model $4$, in which three decoder branches are completely independent without any interaction. The performance gap between Model $3$ and Model $4$ indicates the necessity of information interaction among different types of features. 
\\
\textbf{Multi-modal Filtered Transformer.} Based on Model $3$, we verify the benefits of multi-modal transformer (MMT) and modality-specific filter (MSF) in multi-modal filtered transformer (MFT), as shown in Tab.~\ref{tab:ablation_study_rgbdsod}. 
Model $5$ performs much better than Model $3$, which shows the superiority of multi-modal transformer in the SOD. We conduct a series of experiments about the number of iterations in MMT. The Model $8$ with six MMT iterations achieves the best accuracy among Model $5$ - Model $9$. Then, we equip with the MSF and find that Model $10$  achieves further performance improvement over Model $8$, which verifies the effectiveness of the MSF.
In addition, we replace the MFT module with some convolution layers or general non-local module, and a similar number of parameters and computation are preserved. Compared to Model $11$ and Model $12$, Model $10$ has obvious performance advantage. 
\begin{figure}
	\centering
	\includegraphics[width=\linewidth]{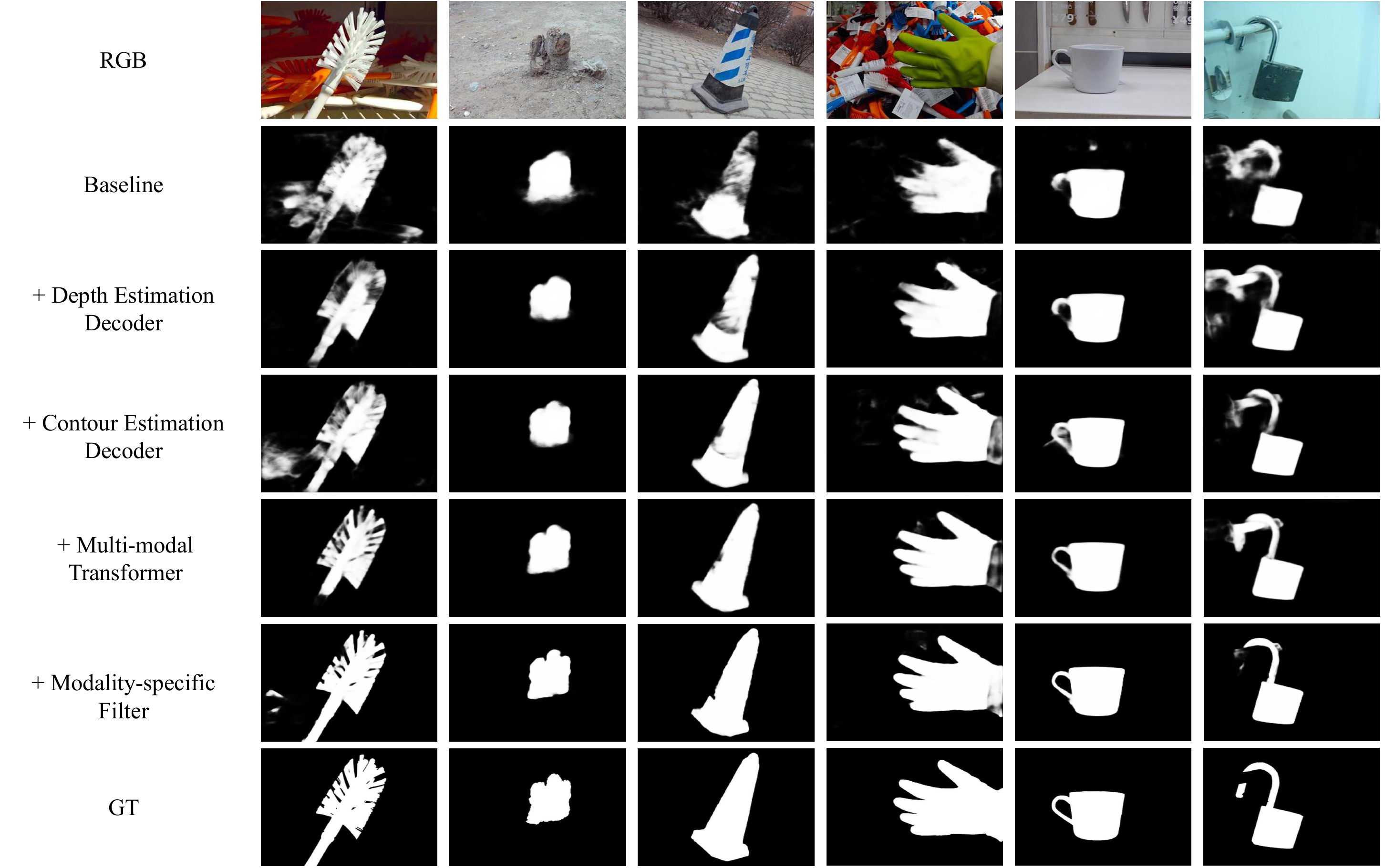}
	\setlength{\abovecaptionskip}{-10pt}
	\caption{Visual results of each component.}
	\label{fig:visual_results_each_component}
	\vspace{-5.5mm}
\end{figure}
\begin{figure}
	\includegraphics[width=\linewidth]{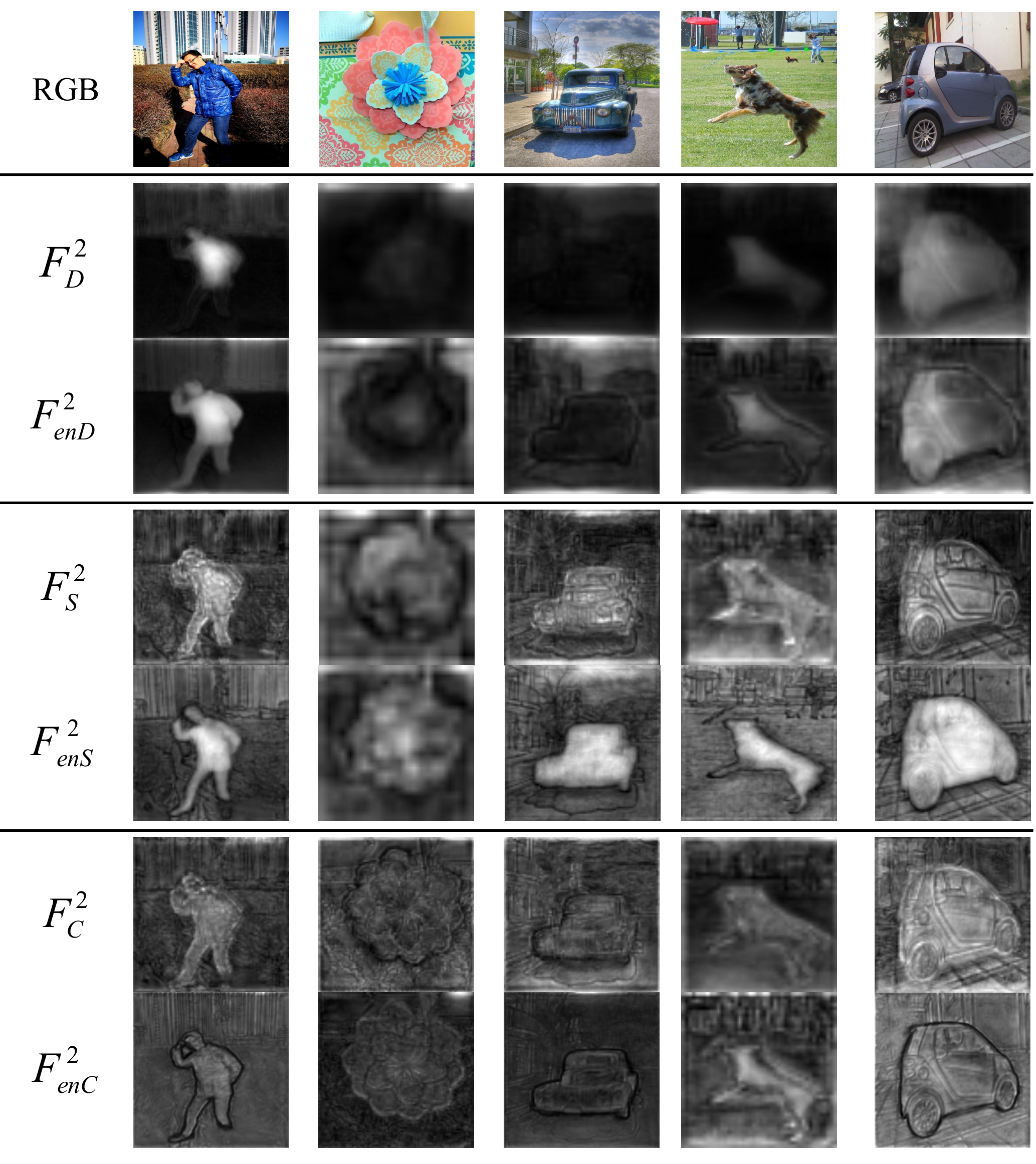}
	\centering
	\setlength{\abovecaptionskip}{-10pt}
	\caption{Visualization of each modality feature in MDB-2. } 
	\label{fig:MDB_visualization}
	 \vspace{-5.5mm}
\end{figure} 
\\
\noindent\textbf{Visual Analyses.} 
Fig.~\ref{fig:visual_results_each_component} shows visual results of the ablation study.  Firstly, we can see that ``+ Depth Estimation Decoder'' model localizes salient object more precisely compared to ``Baseline'' model. Next,  ``+ Contour Estimation Decoder'' model further refine the saliency prediction along the contour. Thirdly,  ``+ Multi-modal Transformer'' model integrates the advantages of three kinds of modalities. Finally,  ``+ Modality-specific Filter'' model significantly refines the details of salient objects.
To intuitively show the effectiveness of the proposed \textit{Squeeze-and-Expand} (SE) structure,  we take MDB-2 as an example to visualize each modality feature before and after SE operation in Fig.~\ref{fig:MDB_visualization}. It can be seen that $F_{enD}^{2}$, $F_{enS}^{2}$ and $F_{enC}^{2}$  all have more precise positions and contours  compared with $F_{D}^{2}$,  $F_{S}^{2}$ and  $F_{c}^{2}$. This  verifies that SE can effectively supplement  complementary information   among different modalities.

\subsection{Ablation studies for Depth and Salient Contour Estimation}\label{sec:Ablation studies for Depth and Contour Estimation}
We evaluate each component for depth estimation and contour prediction.  All the experiments are based on the Res2Net-50 backbone. As shown in Tab.~\ref{tab:ablation_study_depth_estimation} and Tab.~\ref{tab:ablation_study_salient_contour}, Model  $2$ vs.  Model $1$ and Model $3$ vs. Model $2$ show the benefits of the multi-task learning.   The performance gap between Model $3$ and Model $4$  indicates the necessity of  embedding Squeeze-and-Expand structure in the decoder to interact information among different types of features. Model $10$ vs. Model $3$ verifies the benefit of the multi-modal filtered transformer.  Model $10$ has obvious performance advantage over Model $11$ and Model $12$ under a similar number of parameters and comparable computational cost.

\section{More Insightful Analyses}
\label{sec:More_Insightful_Analyses}
\subsection{Why can different tasks promote each other?}
\label{sec:Different_Tasks_Promote_Each_Other}
In this work, the three tasks all require to capture vital position and appearance cues about the objects, and they are complementary. Specifically, depth map contains natural contrast information, which provides useful guidance for saliency and contour positioning. Object contour is one of important attributes of saliency and depth modalities. Accurate contour prediction will improve the quality of depth and saliency estimation. Saliency map provides a fine foreground segmentation and its internal consistency is helpful to avoid the checkerboard pattern in depth estimation.

In a multi-task learning network, the backpropagation acts on multiple task branches in parallel. Different  branches share hidden layer features of multiple scales from the encoder network, and the feature representations used for specific task in these hidden layers are also used by the other tasks. These prompt multiple tasks to be learned together. Based on the multi-task framework, the implicit data augmentation, attention focusing, eavesdropping and regularization guarantee the above-mentioned complementary cues to be fully mined and utilized, thereby making the three pixel-level dense prediction tasks promote each other.

\subsection
{Why high-quality depth can be estimated?}
\label{sec:High-quality_Depth_Estimation}
It is an interesting phenomenon that we obtain high-quality depth predictions under the supervision of poor depth maps. 
The regularization of multi-task learning can balance the performance of different tasks since each task is actually also regularized by the other tasks. If the predicted depth map approaches the poor ground-truth, it will interfere with the predictions of the other two tasks. To resist the negative effect, the multi-task network interactively adjusts and balances each task prediction. Thus, the depth branch absorbs useful information from saliency and contour branches, e.g., the clues about smooth object region and sharp object contour.

Moreover,  due to the uncertainty of noise, the poor depth ground-truth is more difficulty fitted than good depth map. As reflected in~\cite{DIP}, when using the gradient descent to optimize convolutional neural network, it is not sensitive to noise and will be faster and easier to learn to get a more natural-looking image. In other words, CNN offers high impedance to noise. Therefore, our model can easily learn important and effective signals in the depth map during the training process. 

In summary, multi-task learning strategy enables the model to balance the losses and gradients of different tasks. By suppressing noise and mining complementary information from the auxiliary tasks, a good depth map is obtained. 

\section{Positive Impacts}
\label{sec:Positive Impacts}
\begin{figure}
	\includegraphics[width=\linewidth]{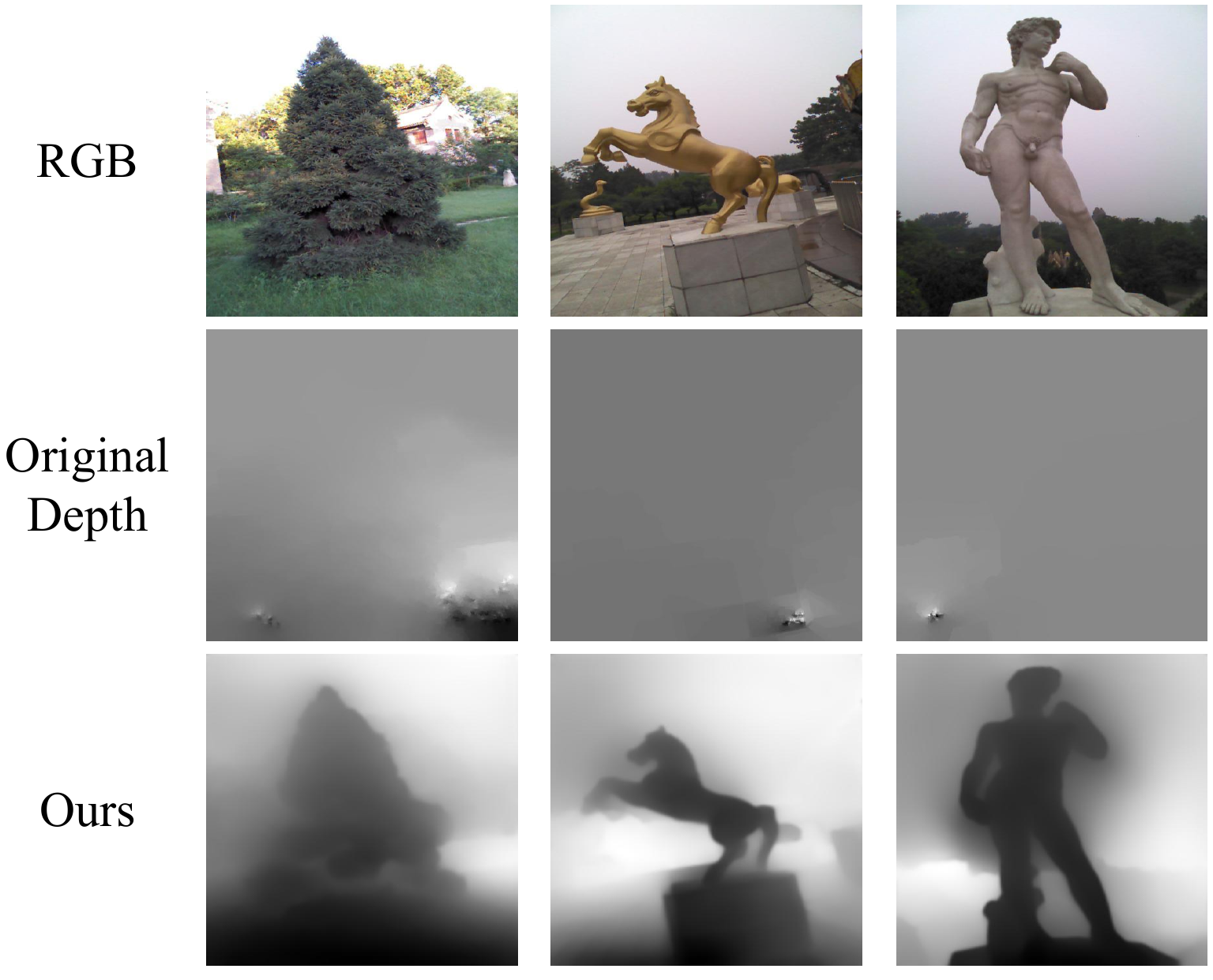}
	\setlength{\abovecaptionskip}{-1pt}
	\centering
	\caption{The case of depth sensor malfunction.} 
	\label{fig:sensor_failure}
\end{figure} 
For the salient object detection task, this work has many positive impacts on the community and practical applications.
\\
\textbf{\textit{First}}, the resulted depth map from our network can help existing depth-based RGB-D SOD methods obtain significant performance gain, as shown in Tab.~\ref{tab:applying_depth_to_rgbdsod}. 
\\
\textbf{\textit{Second}}, once the depth-based method cascades our network and adopt the depth map predicted from our network as the input can help the depth-based network get rid of the dependence on the depth-sensor to have a more flexible applicability. 
\\
\textbf{\textit{Third}}, as shown in Fig.~\ref{fig:sensor_failure}, these examples are from the NLPR~\cite{early_fusion_1} dataset. We can see that the original depth can not capture any depth information, which illustrates the depth sensor do suffer from stability problems, but our approach can avoid such failure predictions. 
\\
\textbf{\textit{Last but not the least}}, we know that one of the important applications of saliency detection is background bokeh. In real smartphones such as the Huawei P40 pro, achieving background bokeh in portrait mode relies on the ToF camera to provide depth information to project bokeh and focus. Different from them, our method not only saves the cost of the ToF camera, but also can provide two modes of AI-assisted background bokeh. They are the extreme mode and the progressive mode with the guidance of saliency map and depth map, respectively.

\section{Limitation and Rethinking}
\label{sec:Limitation and Rethinking}
\begin{figure}
	\includegraphics[width=\linewidth]{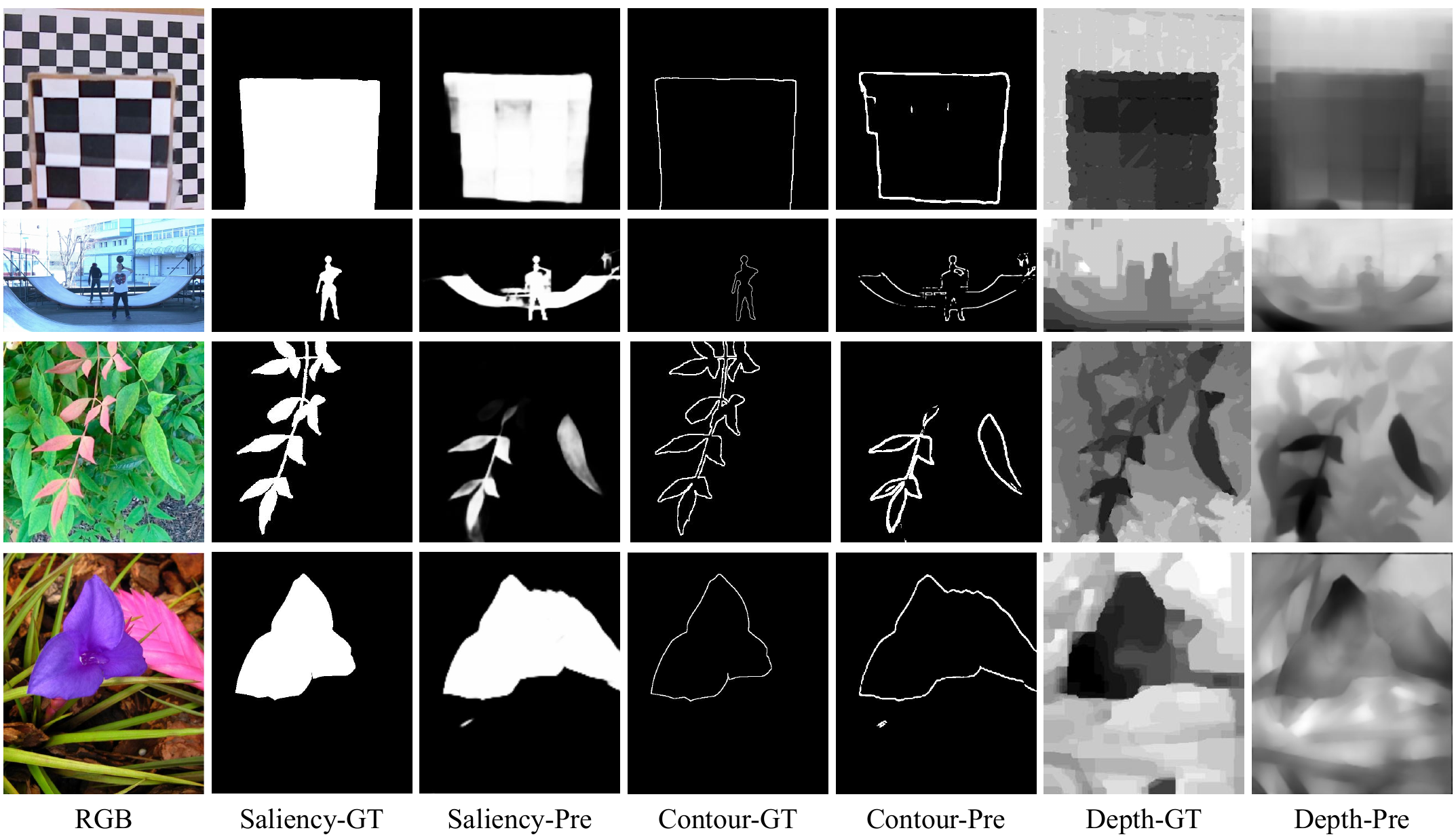}
	\setlength{\abovecaptionskip}{-10pt}
	\centering
	\setlength{\abovecaptionskip}{-8pt}
	\caption{Failure cases.} 
	\label{fig:failure_case}
\end{figure}

In this work, we design three tasks and interact three modal information closely based on our squeeze-and-expand structure. Although we can see a significant improvement in the overall performance under quantitative metrics, there are still some limitations. 
As shown in Fig.~\ref{fig:failure_case}, when one task fails, it possibly affects the predictions of the other two tasks.  In order to avoid this problem,  we give three possible solutions: (1) Construct a dynamic balancing unit to balance the contribution of the features of each modality to the features of other modalities, which can alleviate the negative impact. (2) Set up a task feedback mechanism that scores the confidence of the prediction results through a regression network. According to the scores, the intermediate features generated by each task branch are weighted to refine the prediction results using a recurrent style. (3) Design a balanced loss function to build a controllable sparsity among three tasks. These will be studied in the future.

\section{Conclusion}
In this work, 
we propose a multi-task and multi-modal filtered transformer (MMFT) network for depth-free RGB-D salient object detection. First, we design three related dense prediction tasks: depth estimation, salient object detection and contour prediction. The three modal features rely on the \textit{Squeeze-and-Expand} structure to deeply achieve task-aware information exchanges.  Next, to fully integrate the three kinds of features in high level from the global perspective, we use the transformer technique to achieve multi-modal feature fusion and embed the modality-specific filter to better enhance each modality feature. Through multi-task learning, the depth decoder branch can capture appearance and boundary cues from the other two branches to complete and purify the depth information, thereby predicting a high-quality depth map.
Extensive experiments show that our model performs well on the six RGB-D SOD datasets. By using the predicted depth map, the performance of previous state-of-the-art methods can be consistently improved.

\ifCLASSOPTIONcaptionsoff
  \newpage
\fi

\small{
\bibliography{short_bibtex}
\bibliographystyle{IEEEtran}
}

\end{document}